\documentclass[11pt]{article}

\usepackage[preprint]{acl}

\usepackage{times}
\usepackage{latexsym}

\usepackage[T1]{fontenc}

\usepackage[utf8]{inputenc}

\usepackage{microtype}

\usepackage{inconsolata}
\usepackage{graphicx}
\usepackage{amsmath}
\usepackage{booktabs}
\usepackage{makecell}
\usepackage{amsfonts}
\usepackage{float}
\usepackage{dblfloatfix}
\usepackage{cuted}
\usepackage{caption}
\usepackage[dvipsnames]{xcolor}
\usepackage[most]{tcolorbox}
\usepackage{enumitem}
\usepackage{tabularx}
\usepackage{placeins}
\definecolor{PromptFrame}{RGB}{90,98,112}
\definecolor{PromptTitle}{RGB}{72,78,92}

\title{ContextGuard: Structured Self-Auditing for Context Learning in Language Models}

\author{
 \textbf{Hongbo Jin$^{1}$\thanks{Equal contribution}} \quad
 \textbf{Chi Wang$^{2}$\footnotemark[1]} \quad
   \textbf{Haoran Tang$^{1}$} \quad
   \textbf{Zhongjing Du$^{1}$} \quad \\
  \textbf{Xu Jiang$^{1}$} \quad 
    \textbf{Jingqi Tian$^{3}$} \quad
  \textbf{Qiaoman Zhang$^{1}$} \quad
 \textbf{Jiayu Ding$^{1}$\thanks{Corresponding author}} \\
    $^1$Peking University \quad
    $^2$SCUT \quad
    $^3$Tsinghua University
}

\begin{document}
\maketitle
\begin{abstract}
Recent benchmarks reveal that despite strong reasoning capabilities, large language models (LLMs) still struggle to faithfully apply complex contextual knowledge. These failures are often not wholesale reasoning collapses: in context-rich tasks, models may follow the central reasoning path while missing peripheral, persistent, or format-sensitive requirements. Generic self-refinement is poorly matched to this setting because unconstrained revision may repair one flaw while accidentally breaking already-satisfied constraints. To address this, we introduce ContextGuard, a structured self-auditing framework for protected targeted revision in context learning. ContextGuard decomposes model self-assessment into confirmed constraints, confirmed facts, possibly missed information, and possibly wrong content. Guided by category-conditioned specialist signals, it edits uncertain regions while preserving verified content through explicit protection constraints. Experiments on CL-Bench, a long-context benchmark with densely specified task requirements, show that ContextGuard improves the overall task-solving rate from 9.64\% to 13.85\% (+4.21 percentage points) on Qwen3.5-4B, outperforming generic refinement baselines and reducing failures across format, procedural, calculation, conditional-rule, and style/persona requirements.
\end{abstract}

\section{Introduction}

Recent large language models (LLMs) have achieved remarkable progress in reasoning-intensive domains such as mathematical problem solving, code generation, and agentic planning~\cite{singh2026openaigpt5card}.
Scaling test-time reasoning~\cite{jin2026tir} and reinforcement learning has further improved multi-step inference capabilities~\cite{wang2022self,Guo_2025,jin2026contextcotenhancingcontextlearning,jin2026himachierarchicalmacromicrolearning}.
However, despite these advances, a fundamental capability required by real-world applications remains underdeveloped: the ability to faithfully learn from and apply complex contextual knowledge provided at inference time. In practical deployments, models must increasingly operate in context-rich environments~\cite{bai2024longbench}—such as enterprise rulebooks, legal regulations, and long interaction histories—where success depends on accurately using newly provided information rather than relying solely on static pre-trained knowledge.  

Recent benchmarks such as CL-Bench~\cite{dou2026cl} expose this limitation: even frontier reasoning models perform surprisingly poorly on context-learning tasks, with the strongest models achieving less than 24\% task-solving rate. The challenge is not merely understanding long contexts, but satisfying many contextual requirements simultaneously. CL-Bench operationalizes this through 31,607 binary evaluation rubrics, averaging 16.6 and up to 114 criteria per task; a task is counted as correct only if every associated requirement is satisfied; a task is counted as correct only if every associated requirement is satisfied. As a result, models may solve the primary reasoning objective while still failing due to missed constraints, format rules, role instructions, or contextual exceptions.

\begin{figure*}[t]
    \centering
    \includegraphics[width=\linewidth]{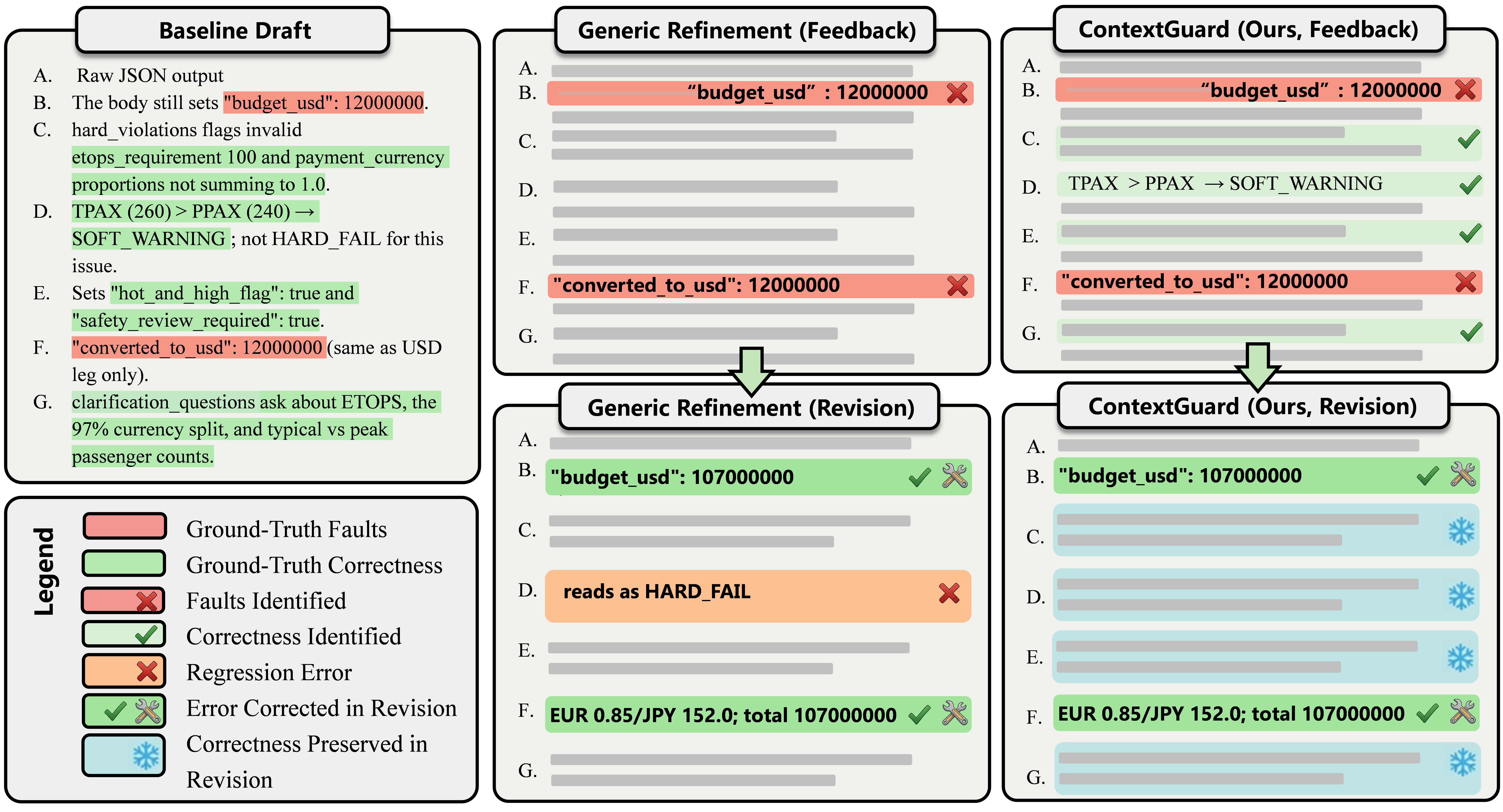}
    \caption{
    Case comparison between vanilla self-refinement and \textbf{ContextGuard} on a representative CL-bench example. 
    The baseline draft contains some \textit{ground-truth faults} (red) and \textit{correct content} (green). 
    Generic refinement successfully fixes some errors but also introduces \textit{revision regression} by incorrectly rewriting previously correct content (orange). 
    In contrast, ContextGuard explicitly distinguishes between \textit{fix targets} and \textit{protected correct regions}, enabling selective correction while preserving verified constraints and facts (blue). 
    }
    \label{fig:comparison}
\end{figure*}

This requirement structure changes what an effective inference-time method must do. In Qwen3.5-4B baseline outputs, 48.3\% of failed tasks miss no more than three criteria, and 72.6\% miss no more than five. These near misses suggest that models often covers the central answer but misses scattered requirements. Generic self-refinement~\cite{madaan2023self, shinn2023reflexion} can partially repair such errors, but unconstrained revision may also damage previously satisfied constraints. Under strict all-requirements evaluation, such revision regressions can erase the benefit of fixing other errors. Existing reasoning-oriented approaches improve deliberative depth~\cite{yao2023tree,Guo_2025}, but do not explicitly preserve already-correct constraints during revision.

In this work, we present ContextGuard, a structured self-auditing framework for protected targeted editing. ContextGuard decomposes generated content into confirmed facts, verified constraints, missed requirements, and potentially wrong reasoning, then revises uncertain regions while anchoring verified content. Category-conditioned specialist signals further target structured failures such as format, workflow, rule-fidelity, and numerical-comparison errors.

We evaluate ContextGuard on CL-Bench across four categories of context-learning tasks: domain knowledge reasoning, rule system application, procedural task execution, and empirical discovery \& simulation. On Qwen3.5-4B, ContextGuard improves the overall task-solving rate from 9.64\% to 13.85\% (+4.21 percentage points), with consistent gains across all task categories. Further analyses show improvements across multiple structured requirement types, including format, procedural coordination, verification, conditional rules, and style/persona constraints.

Our contributions are summarized as follows: 
\begin{itemize}
    \item  We identify a repair-preservation challenge in constraint-dense context learning: effective revision must fix missed requirements without regressing already-satisfied constraints.
    
    \item We propose ContextGuard, a structured self-auditing framework that separates fix targets from protected constraints through epistemic stratification, category-conditioned specialist signals and guarded revision.

    \item We demonstrate substantial improvements on CL-Bench across multiple context-learning categories and provide requirement-level analyses of near-miss behavior, repair/regression dynamics, and diverse contextual requirement types.
\end{itemize}

\section{Related Work}
Unlike traditional \textit{In-Context Learning} which primarily focuses on learning task patterns from a few demonstrations~\cite{brown2020language}, \textit{Context Learning} requires models to to acquire and faithfully apply complex, often novel, contextual knowledge provided at inference time~\cite{dou2026cl}.
This capability is crucial for real-world applications where models must adhere to specific enterprise manuals, legal regulations, or procedural workflows that lie beyond their parametric knowledge.
Recent evaluations on benchmarks like CL-Bench~\cite{dou2026cl} have exposed a significant gap: even state-of-the-art models~\cite{singh2026openaigpt5card,Guo_2025} with strong general reasoning abilities frequently struggle with context-specific constraints.
While long-context LLMs have made strides in information retrieval, successfully solving context-learning tasks requires more than surface-level retrieval~\cite{hsieh2024ruler};
it demands faithful use of the provided constraints during both generation and revision. Our work builds on these insights by introducing a structured auditing mechanism to monitor contextual requirements and constrain revision toward context-faithful outputs.

This view differs from standard long-context retrieval and reasoning-time scaling. Retrieval-oriented methods improve exposure to relevant information, and reasoning-oriented methods increase deliberation, but neither directly addresses the revision objective: preserving satisfied contextual requirements while repairing unsatisfied ones. Further related work is discussed in the appendix~\ref{sec: extended related work}.

\section{Method}

We present \textbf{ContextGuard}, a structured self-auditing framework designed to improve context learning in language models.
Unlike conventional self-refinement methods that perform unconstrained rewriting over the entire response, ContextGuard explicitly models contextual fidelity during inference by separating reliable content from uncertain regions and performing targeted revision under protection constraints.
\begin{figure*}[t]
    \centering
    \includegraphics[width=\linewidth]{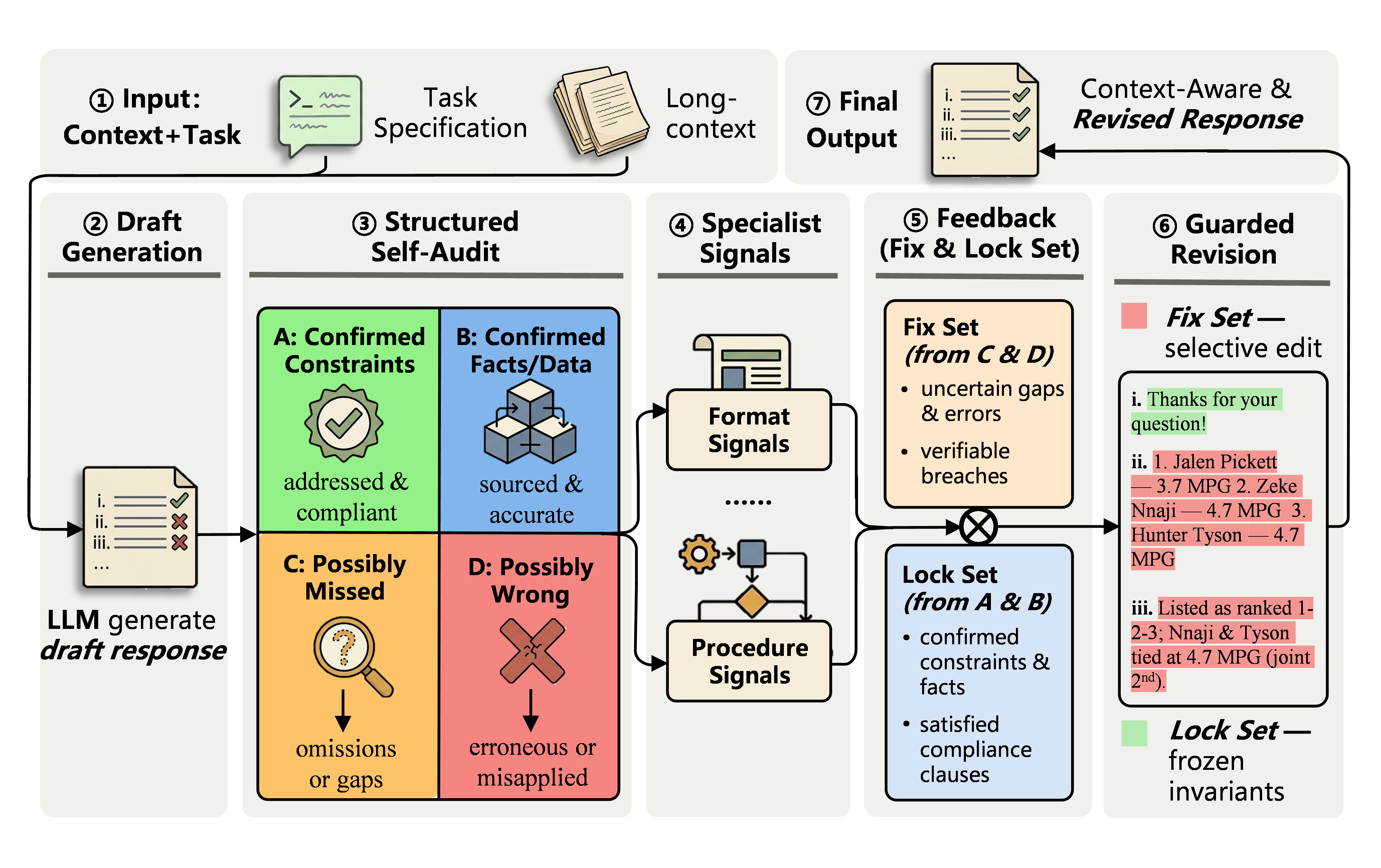}
    \caption{
    Overview of \textbf{ContextGuard}. 
    Given an initial draft generated from the input context and task specification, ContextGuard performs a structured self-audit that partitions model judgments into four epistemic categories: 
    \textbf{(A)} confirmed constraints, 
    \textbf{(B)} confirmed facts/data, 
    \textbf{(C)} possibly missed information, and 
    \textbf{(D)} possibly wrong reasoning or content. 
    Category-conditioned specialist signals are then merged into a \textit{Fix Set} and a \textit{Protection Set}, enabling selective correction while preserving verified constraints and facts.
    }
    \label{fig:pipeline}
\end{figure*}
Figure~\ref{fig:pipeline} illustrates the overall pipeline.
Given a context-learning task, the framework first generates an initial draft, then conducts structured self-auditing to explicitly separate reliable content from uncertain or potentially problematic regions.
Based on the audit results and category-conditioned specialist signals, ContextGuard finally performs protected revision that selectively edits uncertain content while preserving verified information.

\subsection{Overview}

Given a context \(C\) and a task query \(q\), a language model first produces an initial response:
\begin{equation}
y^{(0)} = f_\theta(C, q),
\end{equation}
where \(f_\theta\) denotes the underlying language model. ContextGuard subsequently performs a structured audit over the generated response:
\begin{equation}
\mathcal{A}(y^{(0)}) \rightarrow \{Q_A, Q_B, Q_C, Q_D\},
\end{equation}
where the four subsets \(Q_A, Q_B, Q_C, Q_D\) correspond to different epistemic regions over its own output along two dimensions: correctness and certainty.

For a task type \(t\), ContextGuard may also activate a category-conditioned specialist signal:
\begin{equation}
\mathcal{S}_t(C, q, y^{(0)}) \rightarrow (\mathcal{O}_t, \mathcal{E}_t),
\end{equation}
where \(\mathcal{O}_t\) denotes satisfied specialist requirements and \(\mathcal{E}_t\) denotes detected specialist issues. Depending on the category, \(\mathcal{S}_t\) is implemented either as a separate checker or as specialist criteria integrated into the structured audit.

Based on the auditing results and specialist signals, the framework constructs a structured feedback signal consisting of a fix set \(\mathcal{F}\) containing potentially problematic content and a protection set \(\mathcal{P}\) containing verified information that should remain unchanged. The final response is then generated through constrained revision:
\begin{equation}
y^\ast = \mathcal{R}_{\text{guarded}}(C, q, y^{(0)}, \mathcal{F}, \mathcal{P}),
\end{equation}
where \(\mathcal{R}_{\text{guarded}}\) denotes the protected revision process.
The central idea behind ContextGuard is that effective self-correction requires distinguishing \emph{what should be revised} from \emph{what must be preserved}. Under dense all-requirement evaluation, this distinction is part of the objective rather than a conservative decoding preference: a useful revision should increase the number of satisfied requirements while minimizing regressions on requirements that already pass.

\subsection{Draft Generation}

ContextGuard begins by generating an initial draft conditioned on the original context and user query.
Additionally, we introduce a lightweight reminder augmentation mechanism that explicitly re-emphasizes the original system constraints and final task instruction during draft generation.
Specifically, we append an auxiliary reminder constructed from the original system prompt and task request:
\begin{equation}
r = \texttt{Reminder}(s, q),
\end{equation}
where \(s\) denotes the original system instruction.
The draft generation after augmentation becomes:
\begin{equation}
y^{(0)}_{\text{rem}} = f_\theta(C, q, r).
\end{equation}

This design is motivated by the observation that, in long inputs, global instructions and always-on constraints can become less salient relative to the final task request. By explicitly re-anchoring critical contextual constraints before generation, the model becomes more likely to preserve task requirements throughout the reasoning process. In practice, the reminder-augmented draft consistently provides a stronger revision starting point for downstream self-auditing and protected editing.

\subsection{Structured Self-Auditing}

A key limitation of existing self-refinement methods is that they typically ask models to generically ``check for errors'' without distinguishing reliable content from uncertain regions. As a result, the revision process frequently rewrites already-correct information and introduces unnecessary degradation. To address this issue, ContextGuard introduces \emph{epistemic stratification}, a structured self-auditing mechanism that explicitly decomposes the generated response into different semantic confidence regions before revision.

Given an initial draft \(y^{(0)}\), the model performs structured self-assessment:
\begin{equation}
\mathcal{A}(y^{(0)}) = (Q_A, Q_B, Q_C, Q_D),
\end{equation}
where \(Q_A\) contains confirmed correct reasoning steps and satisfied constraints, \(Q_B\) contains verified contextual data and grounded evidence, \(Q_C\) represents potentially missed constraints or incomplete coverage, and \(Q_D\) contains potentially incorrect reasoning, calculations, or conclusions. The auditing stage outputs a structured JSON representation that explicitly separates trusted regions from uncertain regions. Importantly, \(Q_A\) and \(Q_B\) are treated as protected regions during revision, while \(Q_C\) and \(Q_D\) become candidate regions for targeted correction.

Although the overall auditing framework is unified,
different context-learning categories exhibit distinct failure patterns.
ContextGuard therefore adapts the specialist criteria according to task type.
For domain knowledge reasoning tasks, the audit primarily focuses on format and structural consistency, role and persona adherence and contextual applicability.
Procedural execution tasks additionally require verification of workflow ordering, agent coordination, timing constraints, and gate conditions. Rule-system tasks emphasize rule fidelity, exception handling, terminology consistency, and applicability conditions, while empirical discovery tasks focus on numerical correctness, trend interpretation, evidence grounding, and comparison completeness. For empirical tasks, each issue is further assigned a fine-grained error type such as \texttt{numeric}, \texttt{comparison}, or \texttt{coverage}, enabling more precise downstream revision.

\subsection{Category-Conditioned Specialist Signals}

While structured self-auditing captures general uncertainty patterns, some context-learning failures are highly structured and difficult to comprehensively identify through generic reflection alone.
ContextGuard therefore includes a modular specialist-signal layer for structured contextual requirements. The layer is unified at the interface level: each specialist signal produces satisfied requirements \(\mathcal{O}_t\) and detected issues \(\mathcal{E}_t\), which are then consumed by the same guarded revision mechanism.

For categories whose failures are naturally checkable through explicit structural constraints, such as formatting or workflow order, we instantiate the specialist layer as a separate checker. The format signal checks structural formatting requirements such as section organization, JSON validity, ordering constraints, role consistency, citation format, and forbidden content. The procedural signal examines step ordering, agent routing consistency, gate-check execution, timing constraints, logging completeness, and safety behaviors.

For rule-system and empirical tasks, the most important errors are more tightly coupled with the reasoning content itself. We therefore integrate specialist criteria directly into the structured audit rather than adding a separate verifier call. Rule-system criteria focus on exact rule fidelity, exception handling, numerical conditions, terminology consistency, and applicability boundaries. Empirical criteria focus on numerical correctness, comparison completeness, trend interpretation, unit consistency, and evidence grounding. This design keeps ContextGuard unified while allowing the form of specialist feedback to match the dominant failure pattern of each category.

\subsection{Protected Targeted Revision}

After auditing and specialist signaling, ContextGuard constructs two structured sets  for targeted revision.
The fix set aggregates all potentially problematic regions identified during previous stages:
\begin{equation}
\mathcal{F} = Q_C \cup Q_D \cup \mathcal{E}_t,
\end{equation}
which includes missed contextual constraints, potentially incorrect reasoning, and category-specific specialist issues. Simultaneously, the framework constructs a protection set:
\begin{equation}
\mathcal{P} =
Q_A \cup Q_B
\cup
\mathcal{O}_t,
\end{equation}
which contains all verified content that should remain unchanged during revision.

The final revision stage performs constrained editing conditioned on both the fix set and protection set. The model is instructed to supplement missed constraints, correct only verified problematic regions, and fix explicitly detected violations, while preserving all protected content. This transforms self-refinement from unconstrained rewriting into a structured editing process centered on minimal necessary modification.
The protection set is therefore not merely a safety heuristic. It represents the preservation side of the repair-preservation tradeoff induced by strict all-requirement scoring: already-satisfied constraints should remain satisfied while the model repairs missing or incorrect requirements.

To further stabilize the revision process, we introduce a lightweight revision guard that checks for significant deviation from the original draft. If the revised output exhibits excessive structural or length deviation, ContextGuard discards the revision and falls back to the original draft.
This rollback mechanism further improves revision stability in long-context context-learning settings.

\section{Experiments}

\subsection{Experimental Setup}

\paragraph{Benchmark.}
We evaluate ContextGuard on \textbf{CL-Bench}, a recently proposed benchmark designed to measure context learning capabilities in language models. CL-Bench contains four task categories: \textit{Domain Knowledge Reasoning}, \textit{Procedural Task Execution}, \textit{Rule System Application}, and \textit{Empirical Discovery \& Simulation}. Unlike conventional reasoning benchmarks that primarily evaluate parametric reasoning ability, CL-Bench requires models to faithfully apply newly provided contextual knowledge at inference time.

\paragraph{Evaluation Metric.}
Following the official rubric-based evaluation framework of CL-bench, we report the overall \textbf{task solving rate}, where a task is considered successful only if all associated rubrics are judged as \texttt{yes}. Unlike the original protocol that jointly evaluates all rubrics in a single judgment, we assess each rubric independently using three repeated evaluations with majority voting to improve evaluation stability and reduce rubric-level overestimation.

Beyond overall solving rate, we additionally conduct a rubric-level diagnostic analysis over five requirement types: \textit{format/lexical constraints}, \textit{procedure/agent coordination}, \textit{calculation/verification/standards}, \textit{conditional rules}, and \textit{style/audience/persona}. This taxonomy enables fine-grained analysis of structured contextual requirements beyond aggregate task scores.

\paragraph{Base Models.}
We primarily conduct full comparisons on \textbf{Qwen3.5-4B}, a compact reasoning-oriented language model with substantial headroom on context-learning tasks. Qwen3.5-4B is also the default model for the diagnostic analyses, ablations, and appendix studies unless explicitly stated otherwise. To assess whether the repair-preservation mechanism generalizes beyond this compact setting, we additionally evaluate \textbf{Qwen3.5-9B} as a stronger base model under the same ContextGuard pipeline and judge protocol. All base-model generations use the corresponding thinking mode. Unless otherwise specified, all experiments use greedy decoding with identical inference settings between baseline inference and our framework.

\paragraph{Judge Model.}
Rubric evaluation is performed using \textbf{DeepSeek-R1-Distill-Qwen-32B-AWQ}. Following our evaluation protocol, each rubric is assessed independently with three repeated judgments and majority voting. The same judge configuration is used for all methods.

To evaluate the reliability of the proposed protocol, we repeatedly score the same set of model outputs under identical judge settings. Compared with the original single-pass joint evaluation protocol, our rubric-level repeated evaluation achieves substantially higher consistency,  with agreement rates of approximately 92\% at the rubric level and 97\% at the task level across repeated runs. We further observe over 90\% agreement with the original GPT-5.1-based CL-Bench evaluation pipeline on a sampled subset of tasks, suggesting that the proposed protocol improves evaluation stability while remaining highly consistent with the original benchmark judgments.

\paragraph{Implementation Details.}
ContextGuard operates entirely at inference time without additional training or parameter updates. All auditing, specialist signaling, and revision stages are implemented through prompting. During self-auditing, the model generates structured JSON outputs describing the four epistemic regions. specialist signals are activated sequentially according to task type, either through separate checks or audit-integrated criteria. During revision, the model receives both the fix set and protection set together with explicit editing constraints. Unless otherwise specified, we perform at most one revision round to balance effectiveness, stability, and inference cost.

\subsection{Main Results}

Table~\ref{tab:main_results} reports the overall performance on CL-Bench. ContextGuard consistently improves task-solving rate across all four task categories. On Qwen3.5-4B, our framework improves the average solving rate from 9.64\% to 13.85\%, yielding a substantial gain of +4.21 percentage points. Notably, the improvements are consistent across all four macro-categories, demonstrating that the proposed framework generalizes across diverse context-learning scenarios rather than overfitting to a specific failure pattern.

We further include a stronger-model generalization check using Qwen3.5-9B to evaluate whether ContextGuard remains beneficial when the underlying model is stronger. On Qwen3.5-9B, ContextGuard improves the overall task-solving rate from 10.43\% to 15.80\% (+5.37 pp), with positive gains in every task category, including Domain Knowledge Reasoning (+6.34 pp), Empirical Discovery \& Simulation (+6.03 pp), and Procedural Task Execution (+5.52 pp). We omit Self-Refine for the 9B setting because this additional experiment tests the generality of the ContextGuard pipeline rather than repeating the full baseline suite on every model size.

\begin{table*}[t]
\centering
\setlength{\tabcolsep}{4pt}
\begin{tabular}{llrrrrr}
\toprule
\textbf{Model} & \textbf{Method}
& \textbf{Overall}
& \makecell{\textbf{Domain}\\\textbf{Knowledge}\\\textbf{Reasoning}}
& \makecell{\textbf{Rule System}\\\textbf{Application}}
& \makecell{\textbf{Procedural Task}\\\textbf{Execution}}
& \makecell{\textbf{Empirical}\\\textbf{Discovery \&}\\\textbf{Simulation}} \\
\midrule
Qwen3.5-4B & Baseline & 9.64 & 9.80 & 9.90 & 9.13 & 9.55 \\
Qwen3.5-4B & Self-Refine & 10.48 & 11.76 & 10.42 & 9.34 & 9.05 \\
Qwen3.5-4B & $\Delta_{\text{SelfRefine}}$ & \textcolor{NavyBlue}{+0.84} & \textcolor{NavyBlue}{+1.96} & \textcolor{NavyBlue}{+0.52} & \textcolor{NavyBlue}{+0.21} & \textcolor{red}{-0.50} \\
\textbf{Qwen3.5-4B} & \textbf{ContextGuard} &  \textbf{13.85} & \textbf{14.48} & \textbf{13.07} & \textbf{14.01} & \textbf{13.57} \\
Qwen3.5-4B & $\Delta_{\text{ContextGuard}}$ & \textbf{\textcolor{JungleGreen}{+4.21}} & \textbf{\textcolor{JungleGreen}{+4.68}} & \textbf{\textcolor{JungleGreen}{+3.17}} & \textbf{\textcolor{JungleGreen}{+4.88}} & \textbf{\textcolor{JungleGreen}{+4.02}} \\
\midrule
Qwen3.5-9B & Baseline & 10.43 & 11.31 & 10.25 & 10.62 & 7.54 \\
\textbf{Qwen3.5-9B} & \textbf{ContextGuard} & \textbf{15.80} & \textbf{17.65} & \textbf{14.13} & \textbf{16.14} & \textbf{13.57} \\
Qwen3.5-9B & $\Delta_{\text{ContextGuard}}$ & \textbf{\textcolor{JungleGreen}{+5.37}} & \textbf{\textcolor{JungleGreen}{+6.34}} & \textbf{\textcolor{JungleGreen}{+3.88}} & \textbf{\textcolor{JungleGreen}{+5.52}} & \textbf{\textcolor{JungleGreen}{+6.03}} \\
\bottomrule
\end{tabular}
\caption{Main results and stronger-model generalization on CL-Bench. Task-solving rates (\%) are computed with the official CL-Bench denominators and the same per-requirement majority-vote judge protocol. $\Delta_{\text{SelfRefine}}$ and $\Delta_{\text{ContextGuard}}$ respectively reports Self-Refine's and ContextGuard's absolute gains over the corresponding baselines in percentage points.}
\label{tab:main_results}
\end{table*}

Unless explicitly stated otherwise, the remaining analyses use Qwen3.5-4B, for which we run the full comparison suite. For generic refinement baselines, we use the same initial responses and judge protocol as ContextGuard; Self-Refine performs one round of self-feedback followed by unconstrained revision. Compared with Self-Refine, ContextGuard achieves substantially larger gains because it addresses two weaknesses of generic revision: treating all generated content as freely editable, which can cause destructive rewrites, and lacking explicit checks for structured contextual failures such as format, or rule-application errors.

\subsection{Requirement-Level Failure Analysis}

CL-Bench identifies broad context-learning failures such as ignored context, misused context, and format-following errors. To measure where ContextGuard improves behavior more precisely, we conduct a requirement-level diagnostic analysis before and after applying ContextGuard. Rather than assigning each failed task to a single coarse failure mode, we group failed criteria by the type of contextual requirement they test: \textit{format/lexical constraints}, \textit{procedure/agent coordination}, \textit{calculation/verification/standards}, \textit{conditional rules}, and \textit{style/audience/persona}.

\begin{table}[h]
\centering
\small
\begin{tabular}{lrr}
\toprule
\makecell{Rubric Type} & Baseline & ContextGuard \\
\midrule
Format / lexical & 26.01 & \textbf{23.64} \\
Procedure / agent & 24.65 & \textbf{24.39} \\
Calc. / verify / standards & 31.92 & \textbf{29.71} \\
Conditional rules & 20.40 & \textbf{17.59} \\
Style / audience / persona & 24.78 & \textbf{21.75} \\
\bottomrule
\end{tabular}
\caption{Fine-grained rubric failure analysis on CL-Bench (\%). Failure rates are task-averaged within each rubric type.}
\label{tab:fine_grained_error_analysis}
\end{table}

Table~\ref{tab:fine_grained_error_analysis} shows that the improvements are not limited to surface formatting. ContextGuard also reduces failures related to procedural coordination, verification-oriented requirements, conditional rules, and audience/persona adaptation. These results suggest that many CL-Bench failures arise from structured contextual requirements rather than isolated reasoning errors.

\subsection{Ablation Study}

We conduct ablation studies to quantify the contribution of each major component in ContextGuard. Table~\ref{tab:ablation} reports performance after removing each modules from the full framework.

Removing structured self-auditing leads to the largest performance drop, indicating that epistemic stratification is central to reliable self-correction in context-learning tasks. Removing the protection set (\textit{A+B}) while retaining the fix set (\textit{C+D}) still improves over removing structured auditing entirely, suggesting that targeted correction is beneficial but insufficient on its own. Without explicitly preserving verified content, part of the gain is offset by destructive revisions introduced during editing.

\begin{table}[h]
\centering
\small
\begin{tabular}{lc}
\toprule
Variant & Overall (\%)\\
\midrule
Full Framework & \textbf{13.85} \\
- Structured Self-Audit(A+B) & 11.06 \\
- Structured Self-Audit & 10.53 \\
- Reminder Augmentation & 12.22 \\
- Specialist Signals & 12.84 \\
\bottomrule
\end{tabular}
\caption{Ablation study of ContextGuard on CL-Bench.}
\label{tab:ablation}
\end{table}

Performance further decreases when reminder augmentation or specialist signals are removed, indicating that explicit contextual guidance and targeted failure detection provide complementary benefits beyond structured auditing alone.

\begin{figure*}[t]
    \centering
    \includegraphics[width=\linewidth]{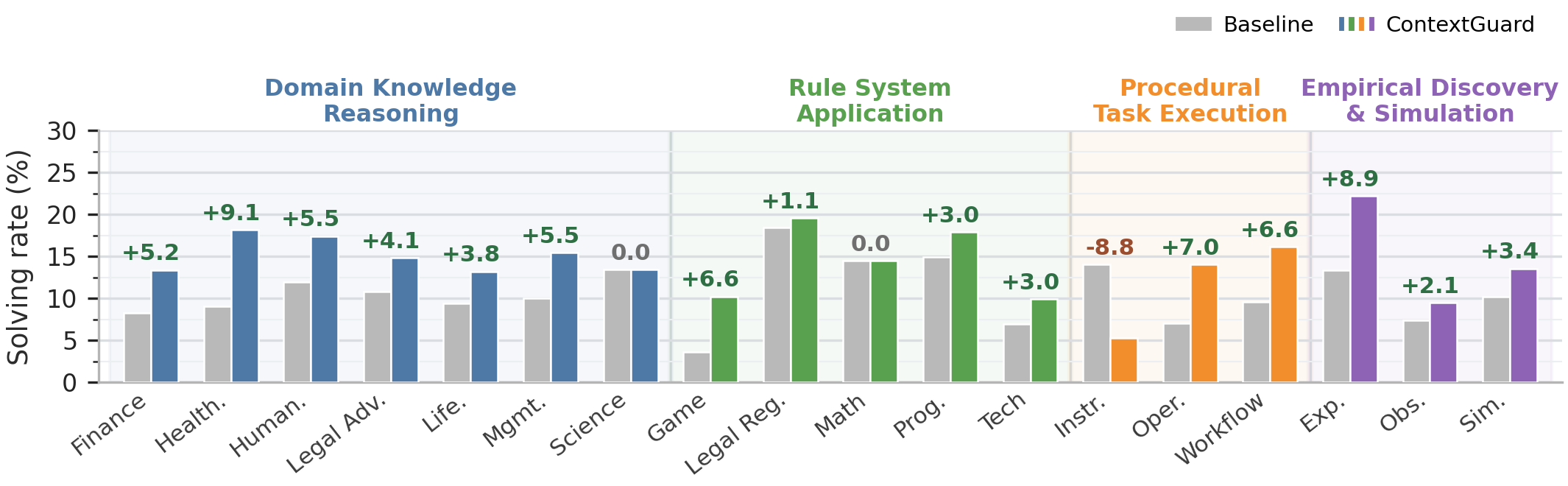}
    \caption{Solving rate (\%) by CL-bench sub-category.
    Sub-categories are ordered and color-grouped by the four benchmark macro-categories.}
    \label{fig:subcategory}
\end{figure*}

\begin{figure}[t]
    \centering
    \includegraphics[width=\linewidth]{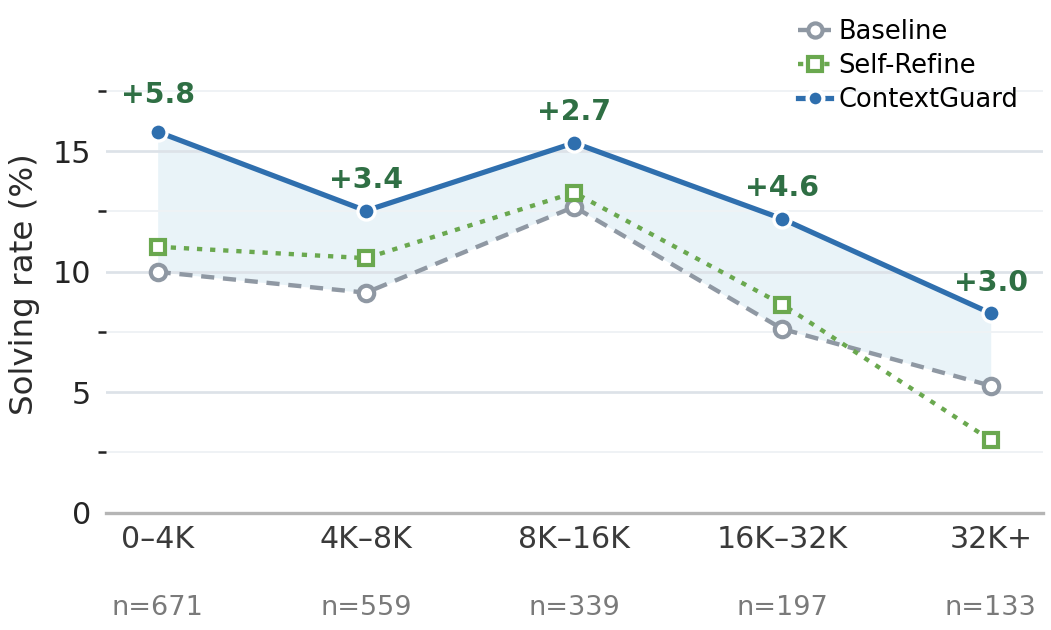}
    \caption{Solving rate (\%) by input-length bin on CL-bench.
    Length is measured as the total number of tokens in all message contents (tiktoken, \texttt{cl100k\_base}).
    Bins are left-closed intervals: 0-4K, 4-8K, 8-16K, 16-32K, 32K+.
    The shaded band and $\Delta$ labels report ContextGuard's absolute gain over the baseline (percentage points) per bin.}
    \label{fig:length}
\end{figure}

Overall, the ablation results suggest that ContextGuard's effectiveness arises from the interaction between epistemic decomposition, protected editing, reminder augmentation, and specialist signaling with each component contributing complementary, non-redundant benefits. In particular, the protection-set ablation connects directly to the revision dynamics above: identifying fix targets is not enough under strict scoring unless the revision process also preserves constraints that already pass.

\subsection{Performance Across Context Lengths}

We also analyze model performance across different context-length ranges.
Figure~\ref{fig:length} compares three inference-time pipelines:  \textbf{Baseline}, one-round generic \textbf{Self-Refine}, and \textbf{ContextGuard}. Across length bins, all methods exhibit a generally downward trend with local fluctuations, reflecting the increasing difficulty of long-context CL tasks.
Nevertheless, ContextGuard consistently outperforms the baseline across all length ranges, maintaining stable gains of +2.7 to +5.8 percentage points.
In contrast, generic Self-Refine provides only marginal improvements in shorter contexts and collapses in the 32K+ bin. This suggests that unconstrained rewriting becomes increasingly unstable on highly constraint-dense inputs. By comparison, ContextGuard still preserves a positive gain in the same regime.
We hypothesize that longer contexts amplify failure modes such as constraint omission, contextual drift, and format inconsistency while also increasing the risk of destructive revisions. These results suggest that structured self-auditing and protected revision become increasingly important as context complexity grows.

\section{Conclusion}

We presented \textbf{ContextGuard}, an inference-time framework for context learning under dense all-requirement evaluation. Instead of treating self-refinement as unconstrained rewriting, ContextGuard separates verified constraints and facts from missed or uncertain content, then performs protected targeted revision with category-conditioned specialist signals. On CL-Bench, this repair-preservation design improves task-solving performance across all four categories and reduces failures in format, procedure, calculation/verification, conditional-rule, and style/persona requirements. Our analyses show that many failures are near misses: models often satisfy most requirements, and gains depend on repairing the remaining errors without regressing already-correct content. These results suggest that reliable context learning requires preservation-aware revision, not only stronger reasoning or longer contexts.

\section*{Limitations}

While ContextGuard proposes an effective structured self-auditing pipeline for context-learning tasks, our experiments focus primarily on the Qwen3.5 series due to computational resource constraints, without extensively covering a broader range of model families and scales.
In addition, all experiments are conducted exclusively on CL-Bench. Although CL-Bench covers diverse contextual reasoning tasks, it remains unclear whether the observed improvements generalize to other context-learning benchmarks with different context sources, task distributions, or evaluation settings.
Future work will evaluate ContextGuard on additional open-source and proprietary models, broader model scales, and other context-learning benchmarks.



\bibliography{custom}

@misc{singh2026openaigpt5card,
      title={OpenAI GPT-5 System Card}, 
      author={Aaditya Singh and Adam Fry and Adam Perelman and Adam Tart and Adi Ganesh and Ahmed El-Kishky and Aidan McLaughlin and Aiden Low and AJ Ostrow and Akhila Ananthram and Akshay Nathan and Alan Luo and Alec Helyar and Aleksander Madry and Aleksandr Efremov and Aleksandra Spyra and Alex Baker-Whitcomb and Alex Beutel and Alex Karpenko and Alex Makelov and Alex Neitz and Alex Wei and Alexandra Barr and Alexandre Kirchmeyer and Alexey Ivanov and Alexi Christakis and Alistair Gillespie and Allison Tam and Ally Bennett and Alvin Wan and Alyssa Huang and Amy McDonald Sandjideh and Amy Yang and Ananya Kumar and Andre Saraiva and Andrea Vallone and Andrei Gheorghe and Andres Garcia Garcia and Andrew Braunstein and Andrew Liu and Andrew Schmidt and Andrey Mereskin and Andrey Mishchenko and Andy Applebaum and Andy Rogerson and Ann Rajan and Annie Wei and Anoop Kotha and Anubha Srivastava and Anushree Agrawal and Arun Vijayvergiya and Ashley Tyra and Ashvin Nair and Avi Nayak and Ben Eggers and Bessie Ji and Beth Hoover and Bill Chen and Blair Chen and Boaz Barak and Borys Minaiev and Botao Hao and Bowen Baker and Brad Lightcap and Brandon McKinzie and Brandon Wang and Brendan Quinn and Brian Fioca and Brian Hsu and Brian Yang and Brian Yu and Brian Zhang and Brittany Brenner and Callie Riggins Zetino and Cameron Raymond and Camillo Lugaresi and Carolina Paz and Cary Hudson and Cedric Whitney and Chak Li and Charles Chen and Charlotte Cole and Chelsea Voss and Chen Ding and Chen Shen and Chengdu Huang and Chris Colby and Chris Hallacy and Chris Koch and Chris Lu and Christina Kaplan and Christina Kim and CJ Minott-Henriques and Cliff Frey and Cody Yu and Coley Czarnecki and Colin Reid and Colin Wei and Cory Decareaux and Cristina Scheau and Cyril Zhang and Cyrus Forbes and Da Tang and Dakota Goldberg and Dan Roberts and Dana Palmie and Daniel Kappler and Daniel Levine and Daniel Wright and Dave Leo and David Lin and David Robinson and Declan Grabb and Derek Chen and Derek Lim and Derek Salama and Dibya Bhattacharjee and Dimitris Tsipras and Dinghua Li and Dingli Yu and DJ Strouse and Drew Williams and Dylan Hunn and Ed Bayes and Edwin Arbus and Ekin Akyurek and Elaine Ya Le and Elana Widmann and Eli Yani and Elizabeth Proehl and Enis Sert and Enoch Cheung and Eri Schwartz and Eric Han and Eric Jiang and Eric Mitchell and Eric Sigler and Eric Wallace and Erik Ritter and Erin Kavanaugh and Evan Mays and Evgenii Nikishin and Fangyuan Li and Felipe Petroski Such and Filipe de Avila Belbute Peres and Filippo Raso and Florent Bekerman and Foivos Tsimpourlas and Fotis Chantzis and Francis Song and Francis Zhang and Gaby Raila and Garrett McGrath and Gary Briggs and Gary Yang and Giambattista Parascandolo and Gildas Chabot and Grace Kim and Grace Zhao and Gregory Valiant and Guillaume Leclerc and Hadi Salman and Hanson Wang and Hao Sheng and Haoming Jiang and Haoyu Wang and Haozhun Jin and Harshit Sikchi and Heather Schmidt and Henry Aspegren and Honglin Chen and Huida Qiu and Hunter Lightman and Ian Covert and Ian Kivlichan and Ian Silber and Ian Sohl and Ibrahim Hammoud and Ignasi Clavera and Ikai Lan and Ilge Akkaya and Ilya Kostrikov and Irina Kofman and Isak Etinger and Ishaan Singal and Jackie Hehir and Jacob Huh and Jacqueline Pan and Jake Wilczynski and Jakub Pachocki and James Lee and James Quinn and Jamie Kiros and Janvi Kalra and Jasmyn Samaroo and Jason Wang and Jason Wolfe and Jay Chen and Jay Wang and Jean Harb and Jeffrey Han and Jeffrey Wang and Jennifer Zhao and Jeremy Chen and Jerene Yang and Jerry Tworek and Jesse Chand and Jessica Landon and Jessica Liang and Ji Lin and Jiancheng Liu and Jianfeng Wang and Jie Tang and Jihan Yin and Joanne Jang and Joel Morris and Joey Flynn and Johannes Ferstad and Johannes Heidecke and John Fishbein and John Hallman and Jonah Grant and Jonathan Chien and Jonathan Gordon and Jongsoo Park and Jordan Liss and Jos Kraaijeveld and Joseph Guay and Joseph Mo and Josh Lawson and Josh McGrath and Joshua Vendrow and Joy Jiao and Julian Lee and Julie Steele and Julie Wang and Junhua Mao and Kai Chen and Kai Hayashi and Kai Xiao and Kamyar Salahi and Kan Wu and Karan Sekhri and Karan Sharma and Karan Singhal and Karen Li and Kenny Nguyen and Keren Gu-Lemberg and Kevin King and Kevin Liu and Kevin Stone and Kevin Yu and Kristen Ying and Kristian Georgiev and Kristie Lim and Kushal Tirumala and Kyle Miller and Lama Ahmad and Larry Lv and Laura Clare and Laurance Fauconnet and Lauren Itow and Lauren Yang and Laurentia Romaniuk and Leah Anise and Lee Byron and Leher Pathak and Leon Maksin and Leyan Lo and Leyton Ho and Li Jing and Liang Wu and Liang Xiong and Lien Mamitsuka and Lin Yang and Lindsay McCallum and Lindsey Held and Liz Bourgeois and Logan Engstrom and Lorenz Kuhn and Louis Feuvrier and Lu Zhang and Lucas Switzer and Lukas Kondraciuk and Lukasz Kaiser and Manas Joglekar and Mandeep Singh and Mandip Shah and Manuka Stratta and Marcus Williams and Mark Chen and Mark Sun and Marselus Cayton and Martin Li and Marvin Zhang and Marwan Aljubeh and Matt Nichols and Matthew Haines and Max Schwarzer and Mayank Gupta and Meghan Shah and Melody Y. Guan and Melody Huang and Meng Dong and Mengqing Wang and Mia Glaese and Micah Carroll and Michael Lampe and Michael Malek and Michael Sharman and Michael Zhang and Michele Wang and Michelle Pokrass and Mihai Florian and Mikhail Pavlov and Miles Wang and Ming Chen and Mingxuan Wang and Minnia Feng and Mo Bavarian and Molly Lin and Moose Abdool and Mostafa Rohaninejad and Nacho Soto and Natalie Staudacher and Natan LaFontaine and Nathan Marwell and Nelson Liu and Nick Preston and Nick Turley and Nicklas Ansman and Nicole Blades and Nikil Pancha and Nikita Mikhaylin and Niko Felix and Nikunj Handa and Nishant Rai and Nitish Keskar and Noam Brown and Ofir Nachum and Oleg Boiko and Oleg Murk and Olivia Watkins and Oona Gleeson and Pamela Mishkin and Patryk Lesiewicz and Paul Baltescu and Pavel Belov and Peter Zhokhov and Philip Pronin and Phillip Guo and Phoebe Thacker and Qi Liu and Qiming Yuan and Qinghua Liu and Rachel Dias and Rachel Puckett and Rahul Arora and Ravi Teja Mullapudi and Raz Gaon and Reah Miyara and Rennie Song and Rishabh Aggarwal and RJ Marsan and Robel Yemiru and Robert Xiong and Rohan Kshirsagar and Rohan Nuttall and Roman Tsiupa and Ronen Eldan and Rose Wang and Roshan James and Roy Ziv and Rui Shu and Ruslan Nigmatullin and Saachi Jain and Saam Talaie and Sam Altman and Sam Arnesen and Sam Toizer and Sam Toyer and Samuel Miserendino and Sandhini Agarwal and Sarah Yoo and Savannah Heon and Scott Ethersmith and Sean Grove and Sean Taylor and Sebastien Bubeck and Sever Banesiu and Shaokyi Amdo and Shengjia Zhao and Sherwin Wu and Shibani Santurkar and Shiyu Zhao and Shraman Ray Chaudhuri and Shreyas Krishnaswamy and Shuaiqi and Xia and Shuyang Cheng and Shyamal Anadkat and Simón Posada Fishman and Simon Tobin and Siyuan Fu and Somay Jain and Song Mei and Sonya Egoian and Spencer Kim and Spug Golden and SQ Mah and Steph Lin and Stephen Imm and Steve Sharpe and Steve Yadlowsky and Sulman Choudhry and Sungwon Eum and Suvansh Sanjeev and Tabarak Khan and Tal Stramer and Tao Wang and Tao Xin and Tarun Gogineni and Taya Christianson and Ted Sanders and Tejal Patwardhan and Thomas Degry and Thomas Shadwell and Tianfu Fu and Tianshi Gao and Timur Garipov and Tina Sriskandarajah and Toki Sherbakov and Tomek Korbak and Tomer Kaftan and Tomo Hiratsuka and Tongzhou Wang and Tony Song and Tony Zhao and Troy Peterson and Val Kharitonov and Victoria Chernova and Vineet Kosaraju and Vishal Kuo and Vitchyr Pong and Vivek Verma and Vlad Petrov and Wanning Jiang and Weixing Zhang and Wenda Zhou and Wenlei Xie and Wenting Zhan and Wes McCabe and Will DePue and Will Ellsworth and Wulfie Bain and Wyatt Thompson and Xiangning Chen and Xiangyu Qi and Xin Xiang and Xinwei Shi and Yann Dubois and Yaodong Yu and Yara Khakbaz and Yifan Wu and Yilei Qian and Yin Tat Lee and Yinbo Chen and Yizhen Zhang and Yizhong Xiong and Yonglong Tian and Young Cha and Yu Bai and Yu Yang and Yuan Yuan and Yuanzhi Li and Yufeng Zhang and Yuguang Yang and Yujia Jin and Yun Jiang and Yunyun Wang and Yushi Wang and Yutian Liu and Zach Stubenvoll and Zehao Dou and Zheng Wu and Zhigang Wang},
      year={2026},
      eprint={2601.03267},
      archivePrefix={arXiv},
      primaryClass={cs.CL},
      url={https://arxiv.org/abs/2601.03267}, 
}

@article{Guo_2025,
   title={DeepSeek-R1 incentivizes reasoning in LLMs through reinforcement learning},
   volume={645},
   ISSN={1476-4687},
   url={http://dx.doi.org/10.1038/s41586-025-09422-z},
   DOI={10.1038/s41586-025-09422-z},
   number={8081},
   journal={Nature},
   publisher={Springer Science and Business Media LLC},
   author={Guo, Daya and Yang, Dejian and Zhang, Haowei and Song, Junxiao and Wang, Peiyi and Zhu, Qihao and Xu, Runxin and Zhang, Ruoyu and Ma, Shirong and Bi, Xiao and Zhang, Xiaokang and Yu, Xingkai and Wu, Yu and Wu, Z. F. and Gou, Zhibin and Shao, Zhihong and Li, Zhuoshu and Gao, Ziyi and Liu, Aixin and Xue, Bing and Wang, Bingxuan and Wu, Bochao and Feng, Bei and Lu, Chengda and Zhao, Chenggang and Deng, Chengqi and Ruan, Chong and Dai, Damai and Chen, Deli and Ji, Dongjie and Li, Erhang and Lin, Fangyun and Dai, Fucong and Luo, Fuli and Hao, Guangbo and Chen, Guanting and Li, Guowei and Zhang, H. and Xu, Hanwei and Ding, Honghui and Gao, Huazuo and Qu, Hui and Li, Hui and Guo, Jianzhong and Li, Jiashi and Chen, Jingchang and Yuan, Jingyang and Tu, Jinhao and Qiu, Junjie and Li, Junlong and Cai, J. L. and Ni, Jiaqi and Liang, Jian and Chen, Jin and Dong, Kai and Hu, Kai and You, Kaichao and Gao, Kaige and Guan, Kang and Huang, Kexin and Yu, Kuai and Wang, Lean and Zhang, Lecong and Zhao, Liang and Wang, Litong and Zhang, Liyue and Xu, Lei and Xia, Leyi and Zhang, Mingchuan and Zhang, Minghua and Tang, Minghui and Zhou, Mingxu and Li, Meng and Wang, Miaojun and Li, Mingming and Tian, Ning and Huang, Panpan and Zhang, Peng and Wang, Qiancheng and Chen, Qinyu and Du, Qiushi and Ge, Ruiqi and Zhang, Ruisong and Pan, Ruizhe and Wang, Runji and Chen, R. J. and Jin, R. L. and Chen, Ruyi and Lu, Shanghao and Zhou, Shangyan and Chen, Shanhuang and Ye, Shengfeng and Wang, Shiyu and Yu, Shuiping and Zhou, Shunfeng and Pan, Shuting and Li, S. S. and Zhou, Shuang and Wu, Shaoqing and Yun, Tao and Pei, Tian and Sun, Tianyu and Wang, T. and Zeng, Wangding and Liu, Wen and Liang, Wenfeng and Gao, Wenjun and Yu, Wenqin and Zhang, Wentao and Xiao, W. L. and An, Wei and Liu, Xiaodong and Wang, Xiaohan and Chen, Xiaokang and Nie, Xiaotao and Cheng, Xin and Liu, Xin and Xie, Xin and Liu, Xingchao and Yang, Xinyu and Li, Xinyuan and Su, Xuecheng and Lin, Xuheng and Li, X. Q. and Jin, Xiangyue and Shen, Xiaojin and Chen, Xiaosha and Sun, Xiaowen and Wang, Xiaoxiang and Song, Xinnan and Zhou, Xinyi and Wang, Xianzu and Shan, Xinxia and Li, Y. K. and Wang, Y. Q. and Wei, Y. X. and Zhang, Yang and Xu, Yanhong and Li, Yao and Zhao, Yao and Sun, Yaofeng and Wang, Yaohui and Yu, Yi and Zhang, Yichao and Shi, Yifan and Xiong, Yiliang and He, Ying and Piao, Yishi and Wang, Yisong and Tan, Yixuan and Ma, Yiyang and Liu, Yiyuan and Guo, Yongqiang and Ou, Yuan and Wang, Yuduan and Gong, Yue and Zou, Yuheng and He, Yujia and Xiong, Yunfan and Luo, Yuxiang and You, Yuxiang and Liu, Yuxuan and Zhou, Yuyang and Zhu, Y. X. and Huang, Yanping and Li, Yaohui and Zheng, Yi and Zhu, Yuchen and Ma, Yunxian and Tang, Ying and Zha, Yukun and Yan, Yuting and Ren, Z. Z. and Ren, Zehui and Sha, Zhangli and Fu, Zhe and Xu, Zhean and Xie, Zhenda and Zhang, Zhengyan and Hao, Zhewen and Ma, Zhicheng and Yan, Zhigang and Wu, Zhiyu and Gu, Zihui and Zhu, Zijia and Liu, Zijun and Li, Zilin and Xie, Ziwei and Song, Ziyang and Pan, Zizheng and Huang, Zhen and Xu, Zhipeng and Zhang, Zhongyu and Zhang, Zhen},
   year={2025},
   month=Sept, pages={633–638} }

@misc{wang2022self,
      title={Self-Consistency Improves Chain of Thought Reasoning in Language Models}, 
      author={Xuezhi Wang and Jason Wei and Dale Schuurmans and Quoc Le and Ed Chi and Sharan Narang and Aakanksha Chowdhery and Denny Zhou},
      year={2023},
      eprint={2203.11171},
      archivePrefix={arXiv},
      primaryClass={cs.CL},
      url={https://arxiv.org/abs/2203.11171}, 
}

@inproceedings{bai2024longbench,
      title={LongBench: A Bilingual, Multitask Benchmark for Long Context Understanding}, 
      author={Yushi Bai and Xin Lv and Jiajie Zhang and Hongchang Lyu and Jiankai Tang and Zhidian Huang and Zhengxiao Du and Xiao Liu and Aohan Zeng and Lei Hou and Yuxiao Dong and Jie Tang and Juanzi Li},
      year={2024},
      eprint={2308.14508},
      archivePrefix={arXiv},
      primaryClass={cs.CL},
      url={https://arxiv.org/abs/2308.14508}, 
}

@misc{dou2026cl,
      title={CL-bench: A Benchmark for Context Learning}, 
      author={Shihan Dou and Ming Zhang and Zhangyue Yin and Chenhao Huang and Yujiong Shen and Junzhe Wang and Jiayi Chen and Yuchen Ni and Junjie Ye and Cheng Zhang and Huaibing Xie and Jianglu Hu and Shaolei Wang and Weichao Wang and Yanling Xiao and Yiting Liu and Zenan Xu and Zhen Guo and Pluto Zhou and Tao Gui and Zuxuan Wu and Xipeng Qiu and Qi Zhang and Xuanjing Huang and Yu-Gang Jiang and Di Wang and Shunyu Yao},
      year={2026},
      eprint={2602.03587},
      archivePrefix={arXiv},
      primaryClass={cs.CL},
      url={https://arxiv.org/abs/2602.03587}, 
}

@inproceedings{yao2023tree,
 author = {Yao, Shunyu and Yu, Dian and Zhao, Jeffrey and Shafran, Izhak and Griffiths, Tom and Cao, Yuan and Narasimhan, Karthik},
 booktitle = {Advances in Neural Information Processing Systems},
 editor = {A. Oh and T. Naumann and A. Globerson and K. Saenko and M. Hardt and S. Levine},
 pages = {11809--11822},
 publisher = {Curran Associates, Inc.},
 title = {Tree of Thoughts: Deliberate Problem Solving with Large Language Models},
 url = {https://proceedings.neurips.cc/paper_files/paper/2023/file/271db9922b8d1f4dd7aaef84ed5ac703-Paper-Conference.pdf},
 volume = {36},
 year = {2023}
}

@inproceedings{madaan2023self,
 author = {Madaan, Aman and Tandon, Niket and Gupta, Prakhar and Hallinan, Skyler and Gao, Luyu and Wiegreffe, Sarah and Alon, Uri and Dziri, Nouha and Prabhumoye, Shrimai and Yang, Yiming and Gupta, Shashank and Majumder, Bodhisattwa Prasad and Hermann, Katherine and Welleck, Sean and Yazdanbakhsh, Amir and Clark, Peter},
 booktitle = {Advances in Neural Information Processing Systems},
 editor = {A. Oh and T. Naumann and A. Globerson and K. Saenko and M. Hardt and S. Levine},
 pages = {46534--46594},
 publisher = {Curran Associates, Inc.},
 title = {Self-Refine: Iterative Refinement with Self-Feedback},
 url = {https://proceedings.neurips.cc/paper_files/paper/2023/file/91edff07232fb1b55a505a9e9f6c0ff3-Paper-Conference.pdf},
 volume = {36},
 year = {2023}
}

@inproceedings{shinn2023reflexion,
 author = {Shinn, Noah and Cassano, Federico and Gopinath, Ashwin and Narasimhan, Karthik and Yao, Shunyu},
 booktitle = {Advances in Neural Information Processing Systems},
 editor = {A. Oh and T. Naumann and A. Globerson and K. Saenko and M. Hardt and S. Levine},
 pages = {8634--8652},
 publisher = {Curran Associates, Inc.},
 title = {Reflexion: language agents with verbal reinforcement learning},
 url = {https://proceedings.neurips.cc/paper_files/paper/2023/file/1b44b878bb782e6954cd888628510e90-Paper-Conference.pdf},
 volume = {36},
 year = {2023}
}

@inproceedings{brown2020language,
 author = {Brown, Tom and Mann, Benjamin and Ryder, Nick and Subbiah, Melanie and Kaplan, Jared D and Dhariwal, Prafulla and Neelakantan, Arvind and Shyam, Pranav and Sastry, Girish and Askell, Amanda and Agarwal, Sandhini and Herbert-Voss, Ariel and Krueger, Gretchen and Henighan, Tom and Child, Rewon and Ramesh, Aditya and Ziegler, Daniel and Wu, Jeffrey and Winter, Clemens and Hesse, Chris and Chen, Mark and Sigler, Eric and Litwin, Mateusz and Gray, Scott and Chess, Benjamin and Clark, Jack and Berner, Christopher and McCandlish, Sam and Radford, Alec and Sutskever, Ilya and Amodei, Dario},
 booktitle = {Advances in Neural Information Processing Systems},
 editor = {H. Larochelle and M. Ranzato and R. Hadsell and M.F. Balcan and H. Lin},
 pages = {1877--1901},
 publisher = {Curran Associates, Inc.},
 title = {Language Models are Few-Shot Learners},
 url = {https://proceedings.neurips.cc/paper_files/paper/2020/file/1457c0d6bfcb4967418bfb8ac142f64a-Paper.pdf},
 volume = {33},
 year = {2020}
}

@misc{hsieh2024ruler,
      title={RULER: What's the Real Context Size of Your Long-Context Language Models?}, 
      author={Cheng-Ping Hsieh and Simeng Sun and Samuel Kriman and Shantanu Acharya and Dima Rekesh and Fei Jia and Yang Zhang and Boris Ginsburg},
      year={2024},
      eprint={2404.06654},
      archivePrefix={arXiv},
      primaryClass={cs.CL},
      url={https://arxiv.org/abs/2404.06654}, 
}

@inproceedings{dhuliawala2024chain,
    title = "Chain-of-Verification Reduces Hallucination in Large Language Models",
    author = "Dhuliawala, Shehzaad  and
      Komeili, Mojtaba  and
      Xu, Jing  and
      Raileanu, Roberta  and
      Li, Xian  and
      Celikyilmaz, Asli  and
      Weston, Jason",
    editor = "Ku, Lun-Wei  and
      Martins, Andre  and
      Srikumar, Vivek",
    booktitle = "Findings of the Association for Computational Linguistics: ACL 2024",
    month = aug,
    year = "2024",
    address = "Bangkok, Thailand",
    publisher = "Association for Computational Linguistics",
    url = "https://aclanthology.org/2024.findings-acl.212/",
    doi = "10.18653/v1/2024.findings-acl.212",
    pages = "3563--3578"
}

@inproceedings{gou2024critic,
 author = {Gou, Zhibin and Shao, Zhihong and Gong, Yeyun and shen, yelong and Yang, Yujiu and Duan, Nan and Chen, Weizhu},
 booktitle = {International Conference on Learning Representations},
 editor = {B. Kim and Y. Yue and S. Chaudhuri and K. Fragkiadaki and M. Khan and Y. Sun},
 pages = {57734--57811},
 title = {CRITIC: Large Language Models Can Self-Correct with Tool-Interactive Critiquing},
 url = {https://proceedings.iclr.cc/paper_files/paper/2024/file/fef126561bbf9d4467dbb8d27334b8fe-Paper-Conference.pdf},
 volume = {2024},
 year = {2024}
}

@inproceedings{huang2024large,
 author = {Huang, Jie and Chen, Xinyun and Mishra, Swaroop and Zheng, Huaixiu Steven and Yu, Adams and Song, Xinying and Zhou, Denny},
 booktitle = {International Conference on Learning Representations},
 editor = {B. Kim and Y. Yue and S. Chaudhuri and K. Fragkiadaki and M. Khan and Y. Sun},
 pages = {32808--32824},
 title = {Large Language Models Cannot Self-Correct Reasoning Yet},
 url = {https://proceedings.iclr.cc/paper_files/paper/2024/file/8b4add8b0aa8749d80a34ca5d941c355-Paper-Conference.pdf},
 volume = {2024},
 year = {2024}
}

@inproceedings{ouyang2022training,
 author = {Ouyang, Long and Wu, Jeffrey and Jiang, Xu and Almeida, Diogo and Wainwright, Carroll and Mishkin, Pamela and Zhang, Chong and Agarwal, Sandhini and Slama, Katarina and Ray, Alex and Schulman, John and Hilton, Jacob and Kelton, Fraser and Miller, Luke and Simens, Maddie and Askell, Amanda and Welinder, Peter and Christiano, Paul F and Leike, Jan and Lowe, Ryan},
 booktitle = {Advances in Neural Information Processing Systems},
 editor = {S. Koyejo and S. Mohamed and A. Agarwal and D. Belgrave and K. Cho and A. Oh},
 pages = {27730--27744},
 publisher = {Curran Associates, Inc.},
 title = {Training language models to follow instructions with human feedback},
 url = {https://proceedings.neurips.cc/paper_files/paper/2022/file/b1efde53be364a73914f58805a001731-Paper-Conference.pdf},
 volume = {35},
 year = {2022}
}

@inproceedings{asai2024self,
 author = {Asai, Akari and Wu, Zeqiu and Wang, Yizhong and Sil, Avi and Hajishirzi, Hannaneh },
 booktitle = {International Conference on Learning Representations},
 editor = {B. Kim and Y. Yue and S. Chaudhuri and K. Fragkiadaki and M. Khan and Y. Sun},
 pages = {9112--9141},
 title = {Self-RAG: Learning to Retrieve, Generate, and Critique through Self-Reflection},
 url = {https://proceedings.iclr.cc/paper_files/paper/2024/file/25f7be9694d7b32d5cc670927b8091e1-Paper-Conference.pdf},
 volume = {2024},
 year = {2024}
}

@inproceedings{lewis2020retrieval,
 author = {Lewis, Patrick and Perez, Ethan and Piktus, Aleksandra and Petroni, Fabio and Karpukhin, Vladimir and Goyal, Naman and K\"{u}ttler, Heinrich and Lewis, Mike and Yih, Wen-tau and Rockt\"{a}schel, Tim and Riedel, Sebastian and Kiela, Douwe},
 booktitle = {Advances in Neural Information Processing Systems},
 editor = {H. Larochelle and M. Ranzato and R. Hadsell and M.F. Balcan and H. Lin},
 pages = {9459--9474},
 publisher = {Curran Associates, Inc.},
 title = {Retrieval-Augmented Generation for Knowledge-Intensive NLP Tasks},
 url = {https://proceedings.neurips.cc/paper_files/paper/2020/file/6b493230205f780e1bc26945df7481e5-Paper.pdf},
 volume = {33},
 year = {2020}
}

@inproceedings{jiang2023llmlingua,
    title = "{LLML}ingua: Compressing Prompts for Accelerated Inference of Large Language Models",
    author = "Jiang, Huiqiang  and
      Wu, Qianhui  and
      Lin, Chin-Yew  and
      Yang, Yuqing  and
      Qiu, Lili",
    editor = "Bouamor, Houda  and
      Pino, Juan  and
      Bali, Kalika",
    booktitle = "Proceedings of the 2023 Conference on Empirical Methods in Natural Language Processing",
    month = dec,
    year = "2023",
    address = "Singapore",
    publisher = "Association for Computational Linguistics",
    url = "https://aclanthology.org/2023.emnlp-main.825/",
    doi = "10.18653/v1/2023.emnlp-main.825",
    pages = "13358--13376"
}

@inproceedings{wei2022chain,
 author = {Wei, Jason and Wang, Xuezhi and Schuurmans, Dale and Bosma, Maarten and ichter, brian and Xia, Fei and Chi, Ed and Le, Quoc V and Zhou, Denny},
 booktitle = {Advances in Neural Information Processing Systems},
 editor = {S. Koyejo and S. Mohamed and A. Agarwal and D. Belgrave and K. Cho and A. Oh},
 pages = {24824--24837},
 publisher = {Curran Associates, Inc.},
 title = {Chain-of-Thought Prompting Elicits Reasoning in Large Language Models},
 url = {https://proceedings.neurips.cc/paper_files/paper/2022/file/9d5609613524ecf4f15af0f7b31abca4-Paper-Conference.pdf},
 volume = {35},
 year = {2022}
}

@inproceedings{zhou2023context,
    title = "Context-faithful Prompting for Large Language Models",
    author = "Zhou, Wenxuan  and
      Zhang, Sheng  and
      Poon, Hoifung  and
      Chen, Muhao",
    editor = "Bouamor, Houda  and
      Pino, Juan  and
      Bali, Kalika",
    booktitle = "Findings of the Association for Computational Linguistics: EMNLP 2023",
    month = dec,
    year = "2023",
    address = "Singapore",
    publisher = "Association for Computational Linguistics",
    url = "https://aclanthology.org/2023.findings-emnlp.968/",
    doi = "10.18653/v1/2023.findings-emnlp.968",
    pages = "14544--14556"
}

@inproceedings{xie2024adaptive,
 author = {Xie, Jian and Zhang, Kai and Chen, Jiangjie and Lou, Renze and Su, Yu},
 booktitle = {International Conference on Learning Representations},
 editor = {B. Kim and Y. Yue and S. Chaudhuri and K. Fragkiadaki and M. Khan and Y. Sun},
 pages = {35623--35646},
 title = {Adaptive Chameleon  or Stubborn Sloth: Revealing the Behavior of Large Language Models in Knowledge Conflicts},
 url = {https://proceedings.iclr.cc/paper_files/paper/2024/file/99261adc8a6356b38bcf999bba9a26dc-Paper-Conference.pdf},
 volume = {2024},
 year = {2024}
}

@inproceedings{wu2025clashevalquantifyingtugofwarllms,
 author = {Wu, Kevin and Wu, Eric and Zou, James},
 booktitle = {Advances in Neural Information Processing Systems},
 doi = {10.52202/079017-1053},
 editor = {A. Globerson and L. Mackey and D. Belgrave and A. Fan and U. Paquet and J. Tomczak and C. Zhang},
 pages = {33402--33422},
 publisher = {Curran Associates, Inc.},
 title = {ClashEval: Quantifying the tug-of-war between an LLM’s internal prior and external evidence},
 url = {https://proceedings.neurips.cc/paper_files/paper/2024/file/3aa291abc426d7a29fb08418c1244177-Paper-Datasets_and_Benchmarks_Track.pdf},
 volume = {37},
 year = {2024}
}

@misc{jin2026contextcotenhancingcontextlearning,
      title={Context-CoT: Enhancing Context Learning via High-Quality Reasoning Synthesis}, 
      author={Hongbo Jin and Mingnan Zhu and Jingqi Tian and Xu Jiang and Zhongjing Du and Haoran Tang and Siyi Xie and Qiaoman Zhang and Jiayu Ding},
      year={2026},
      eprint={2605.25354},
      archivePrefix={arXiv},
      primaryClass={cs.AI},
      url={https://arxiv.org/abs/2605.25354}, 
}

@misc{jin2026himachierarchicalmacromicrolearning,
      title={HiMAC: Hierarchical Macro-Micro Learning for Long-Horizon LLM Agents}, 
      author={Hongbo Jin and Rongpeng Zhu and Jiayu Ding and Guibo Luo and Ge Li},
      year={2026},
      eprint={2603.00977},
      archivePrefix={arXiv},
      primaryClass={cs.AI},
      url={https://arxiv.org/abs/2603.00977}, 
}

@misc{lin2026visdenhancingvideoreasoning,
      title={VISD: Enhancing Video Reasoning via Structured Self-Distillation}, 
      author={Hao Lin and Kunyang Lv and Xu Jiang and Jingqi Tian and Zhongjing Du and Jiayu Ding and Qiaoman Zhang and Hongbo Jin},
      year={2026},
      eprint={2605.06094},
      archivePrefix={arXiv},
      primaryClass={cs.CV},
      url={https://arxiv.org/abs/2605.06094}, 
}

@article{jin2025videocurl,
  title={VideoCuRL: Video Curriculum Reinforcement Learning with Orthogonal Difficulty Decomposition},
  author={Jin, Hongbo and Lin, Kuanwei and Zhang, Wenhao and Jin, Yichen and Li, Ge},
  journal={arXiv preprint arXiv:2601.00887},
  year={2025}
}

@article{jin2026dgpo,
  title={DGPO: Distribution Guided Policy Optimization for Fine Grained Credit Assignment},
  author={Jin, Hongbo and Zhu, Rongpeng and Du, Zhongjing and Jiang, Xu and Tian, Jingqi and Zhang, Qiaoman and Ding, Jiayu},
  journal={arXiv preprint arXiv:2605.03327},
  year={2026}
}

@article{jin2025videomem,
  title={VideoMem: Enhancing Ultra-Long Video Understanding via Adaptive Memory Management},
  author={Jin, Hongbo and Wang, Qingyuan and Zhang, Wenhao and Liu, Yang and Cheng, Sijie},
  journal={arXiv preprint arXiv:2512.04540},
  year={2025}
}

@article{jin2026tir,
  title={TIR-Flow: Active Video Search and Reasoning with Frozen VLMs},
  author={Jin, Hongbo and Xie, Siyi and Ding, Jiayu and Lin, Kuanwei and Li, Ge},
  journal={arXiv preprint arXiv:2601.06176},
  year={2026}
}

\newpage
\appendix

\section{Extended Related Works}
\label{sec: extended related work}

In-context learning (ICL) studies inference-time adaptation from contextual information without parameter updates~\cite{brown2020language}. Context Learning (CL), as instantiated by CL-bench~\cite{dou2026cl}, places stricter requirements on such adaptation by requiring models to internalize and faithfully apply newly provided contextual knowledge under complex constraints. The resulting gap between general inference-time capability and reliable context utilization motivates methods that improve model behavior at inference time, which we review in the following sections.

\medskip

A direct route to improving inference-time capability is to enhance the underlying model through training or post-training. Instruction tuning, reinforcement learning, and retrieval-aware training improve general abilities such as reasoning, instruction following, and knowledge use~\cite{ouyang2022training,Guo_2025,asai2024self,lin2026visdenhancingvideoreasoning,jin2025videocurl,jin2026dgpo}.

However, stronger general capabilities do not necessarily translate to reliable context learning, where success depends on faithfully internalizing and applying newly provided contextual constraints. This limitation motivates inference-time methods that improve behavior without modifying model parameters.

\medskip

Existing non-training approaches improve inference-time behavior without modifying model parameters. These methods can be organized by the failure modes they target in contextual reasoning.

\paragraph{Exposure-related failures.}
Methods such as retrieval-augmented generation, long-context optimization, and prompt compression improve access to relevant information within context~\cite{lewis2020retrieval,liu2024lost,jiang2023llmlingua}. However, improving information exposure does not guarantee faithful use of all contextual constraints.

\paragraph{Reasoning-related failures.}
Methods including Chain-of-Thought, self-consistency, and Tree-of-Thoughts improve inference through additional deliberation~\cite{wei2022chain,wang2022self,yao2023tree}. These approaches strengthen reasoning but do not explicitly enforce adherence to context-specific constraints.

\paragraph{Faithfulness-related failures.}
A third category focuses on grounding model outputs in provided context rather than parametric knowledge. Context-faithful prompting encourages reliance on supplied evidence~\cite{zhou2023context}, while studies on knowledge conflicts and adversarial contexts show that models may still prioritize internal priors over external information~\cite{xie2024adaptive,wu2025clashevalquantifyingtugofwarllms}. These works highlight the challenge of ensuring consistent context usage, especially under conflicting or ambiguous information.

\paragraph{Revision-related failures.}
Self-Refine introduces an iterative self-improvement loop where the model generates an initial answer, receives self-generated feedback, and revises its output~\cite{madaan2023self}. Reflexion extends this idea to agent-like settings by incorporating verbal self-reflection across multiple trials as a form of memory-guided improvement~\cite{shinn2023reflexion,jin2025videomem}. Chain-of-Verification and tool-assisted critiquing further structure the revision process by explicitly separating generation, verification, and final rewriting steps~\cite{dhuliawala2024chain,gou2024critic}. However, despite these designs, self-correction remains unstable, as revision may still degrade previously correct content when feedback signals are incomplete or misaligned~\cite{huang2024large}.

\medskip

Existing inference-time methods address different failure modes in contextual reasoning, but rarely consider how to preserve correct contextual information while performing targeted correction. This limitation becomes critical in context learning settings, where success depends on both faithful constraint use and stable revision.

\medskip

ContextGuard is designed as a unified inference-time framework for such settings. Instead of introducing new training objectives, it restructures inference-time computation through reminder, audit, and protected revision mechanisms to jointly improve constraint awareness and revision robustness.

From a failure-space perspective, ContextGuard targets two limitations identified in prior work: omission or misapplication of contextual constraints, and revision-induced regression over previously correct content. As a result, ContextGuard is better understood as a protected revision framework for context learning, where preserving correct contextual information is treated as equally important as correcting errors.

\section{CL-Bench Details and Strict Requirement Scoring}
\label{sec:app_clbench_details}

\begin{table*}[t]
\centering
\small
\begin{tabular}{lrrrrrr}
\toprule
Category &
\makecell{Contexts} &
\makecell{Tasks} &
\makecell{Criteria} &
\makecell{Avg. Criteria\\per Task} &
\makecell{Max Criteria\\per Task} &
\makecell{Input Length\\Mean / Max} \\
\midrule
Domain Knowledge Reasoning & 190 & 663 & 11,099 & 16.7 & 74 & 8.3K / 60.0K \\
Rule System Application & 140 & 566 & 8,286 & 14.6 & 75 & 12.2K / 62.2K \\
Procedural Task Execution & 100 & 471 & 9,486 & 20.1 & 59 & 8.5K / 58.5K \\
Empirical Discovery \& Simulation & 70 & 199 & 2,736 & 13.7 & 114 & 16.7K / 65.0K \\
\midrule
Total & 500 & 1,899 & 31,607 & 16.6 & 114 & 10.4K / 65.0K \\
\bottomrule
\end{tabular}
\caption{Statistics of CL-Bench~\cite{dou2026cl}. We use ``criteria'' to refer to the benchmark's binary verification requirements. Input length includes the system prompt, context, and task specification.}
\label{tab:clbench_statistics}
\end{table*}

CL-Bench~\cite{dou2026cl} is designed to evaluate whether language models can acquire and apply newly provided contextual knowledge at inference time. Unlike conventional reasoning benchmarks that often evaluate a single final answer, CL-Bench combines long expert-authored contexts, multiple task categories, sequential dependencies, and dense binary requirement checking. This makes it a useful testbed for studying context learning as a strict requirement-satisfaction problem.

Table~\ref{tab:clbench_statistics} summarizes the benchmark statistics. CL-Bench contains 500 contexts, 1,899 tasks, and 31,607 binary verification criteria across four categories and 18 subcategories. Inputs are long, with an average input length of 10.4K tokens and a maximum of 65K tokens. The benchmark is also densely constrained: each task contains 16.6 criteria on average, and some tasks contain up to 114 criteria. In addition, 51.1\% of tasks are sequential, meaning that later turns may depend on information or decisions from earlier turns.

\paragraph{Strict conjunctive scoring.}
CL-Bench evaluates each task using a set of binary criteria. Let task \(i\) contain \(m_i\) criteria, and let \(z_{ij} \in \{0,1\}\) denote whether the model output satisfies criterion \(j\). A task is counted as solved only when all associated criteria are satisfied:
\begin{equation}
s_i = \prod_{j=1}^{m_i} z_{ij}.
\end{equation}
The overall task-solving rate is then
\begin{equation}
\mathrm{TSR} = \frac{1}{N}\sum_{i=1}^{N} s_i,
\end{equation}
where \(N\) is the number of tasks. This strict conjunctive metric differs from partial-credit evaluation: an answer may satisfy most requirements but still receive zero task-level credit if it misses a single required constraint, format rule, exception, calculation detail, or procedural condition.

This scoring rule is central to our motivation. In dense context-learning settings, a model may correctly follow the main reasoning path while failing due to a small number of scattered requirements. Conversely, a revision step may fix an initially failed requirement but accidentally break another requirement that was already satisfied. Therefore, effective refinement must optimize not only for error correction, but also for preservation of already-correct contextual content.

\paragraph{Context dependence.}
CL-Bench is constructed to require the supplied context rather than external retrieval or parametric knowledge alone. The benchmark includes expert-authored contexts containing fictional, modified, niche, or recently emerging information. The original CL-Bench paper reports that when contexts are removed in a sampled ablation, GPT-5.1 (High) drops to 0.9\% task-solving rate on 1,000 sampled tasks~\cite{dou2026cl}. This supports treating CL-Bench as a context-learning benchmark: success depends on using the provided context as the governing source of information.

\paragraph{Benchmark difficulty.}
The original CL-Bench evaluation reports low task-solving rates even for strong frontier models. Across ten evaluated models, the average task-solving rate is 17.2\%, and the strongest reported model, GPT-5.1 (High), reaches only 23.7\% overall~\cite{dou2026cl}. These numbers should be interpreted as evidence of benchmark difficulty rather than directly comparable baselines for our experiments, since our evaluation uses a different judge model and a per-criterion repeated-judgment protocol.

\section{Evaluation Protocol and Judge Stability}
\label{sec:app_eval_protocol}

\paragraph{Per-requirement judging.}
The original CL-Bench evaluation presents all requirements of a task to the judge in a single prompt. In our evaluation, we instead score each requirement independently. This reduces judge attention burden when a task contains many fine-grained requirements and makes the evaluation less sensitive to prompt-position effects among requirements.

For task \(i\) with requirements \(\{r_{ij}\}_{j=1}^{m_i}\), each requirement receives a binary label \(\hat{z}_{ij}\in\{0,1\}\). A task is considered solved only if all requirements are satisfied:
\begin{equation}
\hat{s}_i = \prod_{j=1}^{m_i} \hat{z}_{ij}.
\end{equation}

\paragraph{Single vote vs. majority vote.}
A single LLM judge call can be noisy, especially for long-context answers and partially satisfied requirements. We therefore evaluate each requirement multiple times and aggregate the labels by majority voting:
\begin{equation}
\hat{z}_{ij} =
\mathbb{I}\left[\sum_{k=1}^{K} v_{ij}^{(k)} > \frac{K}{2}\right],
\end{equation}
where \(v_{ij}^{(k)}\in\{0,1\}\) is the \(k\)-th judge vote for requirement \(j\), and \(K\) is the number of repeated judge votes. In all experiments reported in this paper, we use \(K=3\).

\begin{table}[h]
\centering
\small
\begin{tabular}{lrrr}
\toprule
Level & Instances & \makecell{Single-vote\\agreement} & \makecell{Majority-vote\\agreement} \\
\midrule
Requirement & 62{,}308 & 88.29 & 94.14 \\
Task & 3{,}720 & 93.46 & 96.73 \\
\bottomrule
\end{tabular}
\caption{
Judge stability (\%) computed from stored three-vote traces over baseline and ContextGuard outputs. Single-vote agreement measures consistency among repeated individual judge calls. Majority-vote agreement measures agreement between individual votes and the final majority label.
}
\label{tab:judge_vote_stability}
\end{table}

Table~\ref{tab:judge_vote_stability} shows that single judge calls have non-negligible variance. At the requirement level, repeated single votes agree 88.29\% of the time, while agreement with the majority label reaches 94.14\%. At the task level, the corresponding numbers are 93.46\% and 96.73\%. We therefore use per-requirement majority-vote judging with \(K=3\) for all reported results.

\begin{table*}[h]
\centering
\small
\begin{tabular}{p{0.96\linewidth}}
\toprule
\textbf{Algorithm 1: ContextGuard Inference} \\
\midrule
\textbf{Input:} context \(C\), task query \(q\), system instruction \(s\), task category \(t\), model \(f_\theta\) \\
\textbf{Output:} final answer \(y^\ast\) \\
\\[-0.8em]
1. Construct reminder \(r \leftarrow \texttt{Reminder}(s,q)\). \\
2. Generate reminder-augmented draft \(y^{(0)} \leftarrow f_\theta(C,q,r)\). \\
3. Run structured self-audit:
\[
(Q_A,Q_B,Q_C,Q_D) \leftarrow \mathcal{A}(C,q,y^{(0)}).
\]
4. Run category-conditioned specialist signal:
\[
(\mathcal{O}_t,\mathcal{E}_t) \leftarrow \mathcal{S}_t(C,q,y^{(0)}).
\]
5. Construct fix set:
\[
\mathcal{F} \leftarrow Q_C \cup Q_D \cup \mathcal{E}_t.
\]
6. Construct protection set:
\[
\mathcal{P} \leftarrow Q_A \cup Q_B \cup \mathcal{O}_t.
\]
7. \textbf{if} \(\mathcal{F}=\emptyset\) \textbf{then} return \(y^\ast \leftarrow y^{(0)}\). \\
8. Generate protected revision:
\[
\tilde{y} \leftarrow \mathcal{R}_{\mathrm{guarded}}(C,q,y^{(0)},\mathcal{F},\mathcal{P}).
\]
9. Apply revision guard:
\[
g \leftarrow \texttt{RevisionGuard}(\tilde{y},y^{(0)},\mathcal{P}).
\]
10. \textbf{if} \(g=\texttt{pass}\) \textbf{then} return \(y^\ast \leftarrow \tilde{y}\). \\
11. \textbf{else} return \(y^\ast \leftarrow y^{(0)}\). \\
\bottomrule
\end{tabular}
\caption{Pseudocode of ContextGuard inference. The model first identifies fix targets and protected content, then performs one guarded revision round with fallback to the original draft when revision appears destructive.}
\label{tab:contextguard_pseudocode}
\end{table*}

\section{Full ContextGuard Algorithm}
\label{sec:app_algorithm}

Table~\ref{tab:contextguard_pseudocode} gives the complete inference-time procedure of ContextGuard, and Table~\ref{tab:specialist_signals} summarizes how category-conditioned specialist signals are normalized into the same fix-set and lock-set interface. The framework operates on a context \(C\), task query \(q\), optional system instruction \(s\), and task category \(t\). It performs at most one protected revision round and does not update model parameters.

\paragraph{Single-round design.}
All experiments in this paper use a single revision round. This keeps inference cost controlled and avoids compounding revision errors across multiple rounds. The goal is not to repeatedly rewrite the answer, but to perform one targeted repair while preserving already-satisfied contextual requirements.

\paragraph{Revision guard.}
The revision guard is a lightweight fallback mechanism applied after protected revision. It is designed to detect destructive edits rather than to re-grade the answer. In our implementation, the guard checks whether the revised answer shows substantial structural or informational degradation relative to the draft. This includes three main cases.

\begin{itemize}
    \item \textbf{Excessive shortening.} If the revised answer is much shorter than the draft, the revision may have deleted required reasoning, supporting evidence, or output fields. We therefore reject revisions whose length falls below a preset fraction of the original draft length.
    
    \item \textbf{Protected-content loss.} If the revision appears to remove or contradict content listed in the protection set \(\mathcal{P}\), it is treated as unsafe. The protection set contains confirmed constraints, verified facts, and satisfied specialist requirements that should remain stable during editing.
    
    \item \textbf{Structural degradation.} If the draft follows an apparent required structure, such as a list, JSON object, table, or sectioned response, but the revision collapses or substantially alters that structure without a corresponding fix target, the revision is considered potentially destructive.
\end{itemize}

If any guard condition is triggered, ContextGuard discards the revised answer and returns the original draft. This conservative fallback is useful in strict context-learning evaluation: a revision that fixes one missed requirement but deletes an already-correct requirement can still fail the task. The guard therefore complements the protection set by preventing high-risk revisions from replacing a safer draft.

\begin{table*}[t]
\centering
\small
\begin{tabular}{p{0.17\linewidth}p{0.18\linewidth}p{0.27\linewidth}p{0.30\linewidth}}
\toprule
Category & Specialist Signal & Main Checks & Revision Use \\
\midrule

Domain Knowledge Reasoning &
Format and contextual applicability signal &
Checks whether the draft follows required structure, output format, ordering constraints, role/persona constraints, forbidden content constraints, and whether domain-specific claims are grounded in the provided context. &
Format or applicability violations are added to \(\mathcal{E}_t\). Satisfied structural and contextual requirements are added to \(\mathcal{O}_t\) so that revision does not damage already-correct domain-specific content. \\

\midrule

Procedural Task Execution &
Workflow and procedure signal &
Checks step ordering, agent routing, dependency handling, gate conditions, timing constraints, logging requirements, safety behavior, and required escalation or refusal paths. &
Detected workflow errors are added to \(\mathcal{E}_t\). Correctly executed steps, valid routing decisions, and satisfied safety constraints are added to \(\mathcal{O}_t\) to prevent revision from changing an already-valid procedure. \\

\midrule

Rule System Application &
Rule-fidelity signal &
Checks exact rule application, exception handling, numerical or symbolic conditions, terminology consistency, applicability boundaries, and whether the draft invents unsupported rules. &
Misapplied, omitted, or invented rules are added to \(\mathcal{E}_t\). Correctly applied rules and satisfied exceptions are added to \(\mathcal{O}_t\), protecting them during later edits. \\

\midrule

Empirical Discovery \& Simulation &
Empirical consistency signal &
Checks numerical correctness, comparison completeness, unit consistency, trend interpretation, evidence grounding, coverage of requested experimental conditions, and whether claims are supported by the provided data. &
Numerical, comparison, coverage, or evidence-grounding issues are added to \(\mathcal{E}_t\). Verified computations, grounded observations, and correctly stated comparisons are added to \(\mathcal{O}_t\). \\

\bottomrule
\end{tabular}
\caption{
Category-conditioned specialist signals used by ContextGuard. Each signal maps category-specific requirement checks into the same fix-set and lock-set interface used by protected revision.
}
\label{tab:specialist_signals}
\end{table*}

\section{Category-Conditioned Specialist Signals}
\label{sec:app_specialist_signals}

Table~\ref{tab:specialist_signals} provides the category-level checks used by the specialist layer; this appendix subsection explains the common interface behind that table. Given the context \(C\), query \(q\), draft answer \(y^{(0)}\), and task category \(t\), the specialist layer produces satisfied category-specific requirements and detected issues:
\begin{equation}
\mathcal{S}_t(C,q,y^{(0)}) \rightarrow (\mathcal{O}_t,\mathcal{E}_t).
\end{equation}
The detected issues \(\mathcal{E}_t\) are added to the fix set, while the satisfied requirements \(\mathcal{O}_t\) are added to the protection set:
\begin{equation}
\mathcal{F}=Q_C \cup Q_D \cup \mathcal{E}_t,
\quad
\mathcal{P}=Q_A \cup Q_B \cup \mathcal{O}_t.
\end{equation}
Thus, all specialist signals serve the same purpose: they convert category-specific requirement checks into revision guidance.

\paragraph{Category conditioning.}
The category \(t\) is the benchmark context category, not a task outcome label. It is used only to select the appropriate verification emphasis for the draft. This avoids applying the same generic checklist to qualitatively different tasks, such as procedural workflows and empirical data interpretation.

\paragraph{Separate checkers and audit-integrated signals.}
The specialist layer can be implemented in two forms. For requirements with explicit structural patterns, such as formatting or workflow order, ContextGuard uses separate checker prompts that produce explicit satisfied and failed items. For requirements that are tightly coupled with reasoning content, such as rule interpretation or empirical comparison, the specialist criteria can be integrated into the structured self-audit prompt. In both cases, the output is normalized into the same \((\mathcal{O}_t,\mathcal{E}_t)\) interface before revision.

\paragraph{Why specialist signals are needed.}
Generic reflection often identifies high-level mistakes but misses category-specific failure modes. For example, a procedural answer may appear logically plausible while violating a required gate condition; an empirical answer may state the correct trend but omit a required comparison; a rule-system answer may reach the right conclusion while applying an exception incorrectly. The specialist layer makes these failure modes explicit before revision.

\paragraph{Interaction with protected revision.}
Specialist signals are not used only to find errors. They also identify category-specific content that should be preserved. This is important because many drafts are partially correct: they may satisfy the required format, follow several workflow steps, or apply some rules correctly while failing elsewhere. By placing satisfied specialist requirements into \(\mathcal{O}_t\), ContextGuard prevents the revision stage from treating the entire answer as freely editable text.

\section{Prompt Templates}
\label{sec:app_prompt_templates}

Tables~\ref{tab:prompt_templates_1}, \ref{tab:prompt_templates_2}, and~\ref{tab:prompt_templates_3} provide the normalized prompt templates used by ContextGuard. Table~\ref{tab:prompt_templates_1} covers reminder-augmented drafting, structured self-auditing, and format compliance checking; Table~\ref{tab:prompt_templates_2} covers procedure, rule-fidelity, and empirical specialist signals; and Table~\ref{tab:prompt_templates_3} covers protected revision and requirement-level judging. Full task contexts are omitted and replaced with bracketed placeholders. The exact wording may vary slightly across categories, but all prompts follow the same interface: issue items are added to the fix set and confirmed items are added to the protection set.

\clearpage
\begin{table*}[t]
\centering
\small
\setlength{\tabcolsep}{6pt}
\renewcommand{\arraystretch}{1.12}
\begin{tabular}{p{0.20\linewidth}p{0.74\linewidth}}
\toprule
Prompt & Template \\
\midrule

\makecell[l]{Prompt 1:\\Reminder-\\Augmented Draft\\Generation}
&
{\ttfamily\small
[ORIGINAL CONTEXT AND CONVERSATION]

\medskip
[REMINDER]

\medskip
Before answering, re-read and follow all instructions in the system/context message, including constraints, role requirements, format requirements, and task-specific rules.

\medskip
The final task request is:

\medskip
[FINAL TASK REQUEST]

\medskip
Now produce the answer to the final task. Use only the provided context and satisfy all stated requirements.
}
\\

\midrule

\makecell[l]{Prompt 2:\\Structured\\Self-Audit}
&
{\ttfamily\small
You are auditing your previous answer for a context-learning task.

\medskip
[CONVERSATION / CONTEXT]

\medskip
[ORIGINAL TASK]

\medskip
[PREVIOUS ANSWER]

\medskip
Carefully compare the previous answer against the provided context, rules, constraints, data, and final task request.

\medskip
Output a JSON object with four fields:

\medskip
\{
  ``confirmed\_correct'': [
    ``Specific constraints, reasoning steps, or output parts that are correct and should be preserved.''
  ],

  ``confirmed\_data'': [
    ``Specific contextual facts, values, rules, or evidence that were used correctly.''
  ],

  ``possibly\_missed'': [
    ``Constraints, facts, requirements, cases, exceptions, or output elements that may be missing or insufficiently covered.''
  ],

  ``possibly\_wrong'': [
    ``Reasoning steps, calculations, rule applications, conclusions, or answer parts that may be incorrect.''
  ]
\}

\medskip
Be specific. Identify concrete items that can guide revision.
}
\\

\midrule

\makecell[l]{Prompt 3:\\Format\\Compliance Signal}
&
{\ttfamily\small
You are a format compliance checker. Your only job is to check whether the response follows the format, structure, and presentation rules from the conversation.

\medskip
[CONVERSATION / CONTEXT]

\medskip
[PREVIOUS ANSWER]

\medskip
Extract every constraint about format, structure, or presentation, including:

\medskip
- output structure, such as bullets, numbered lists, tables, sections, JSON, XML, or schemas;

- required ordering or grouping;

- required labels, headings, field names, or tags;

- typography or presentation rules;

- length, inclusion, exclusion, or forbidden-content constraints.

\medskip
Output a JSON object:

\medskip
\{
  ``format\_ok'': [
    ``Requirement + how the response satisfies it.''
  ],

  ``format\_fail'': [
    ``Requirement + how the response violates it.''
  ]
\}
}
\\

\bottomrule
\end{tabular}
\caption{Prompt templates for reminder-augmented drafting, structured self-auditing, and format compliance checking.}
\label{tab:prompt_templates_1}
\end{table*}

\clearpage

\begin{table*}[t]
\centering
\small
\setlength{\tabcolsep}{6pt}
\renewcommand{\arraystretch}{1.12}
\begin{tabular}{p{0.20\linewidth}p{0.74\linewidth}}
\toprule
Prompt & Template \\
\midrule

\makecell[l]{Prompt 4:\\Procedure\\Compliance Signal}
&
{\ttfamily\small
You are a procedure compliance checker for a procedural task execution scenario. Your job is to verify workflow correctness, not general writing quality.

\medskip
[CONVERSATION / CONTEXT]

\medskip
[PREVIOUS ANSWER]

\medskip
Check the response against all procedural requirements, including:

\medskip
- required step sequence;

- agent routing and handoff rules;

- dependency handling;

- gate checks and threshold conditions;

- timing constraints;

- logging, reporting, or evidence-recording requirements;

- safety, refusal, or escalation rules.

\medskip
Output a JSON object:

\medskip
\{
  ``proc\_ok'': [
    ``Procedural requirement + how the response satisfies it.''
  ],

  ``proc\_fail'': [
    ``Procedural requirement + how the response violates it.''
  ]
\}
}
\\

\midrule

\makecell[l]{Prompt 5:\\Rule-Fidelity\\Signal}
&
{\ttfamily\small
You are a rule-fidelity checker. The task depends on faithfully applying the provided rulebook or rule system.

\medskip
[CONVERSATION / RULEBOOK]

\medskip
[PREVIOUS ANSWER]

\medskip
Check whether the response correctly applies the provided rules. Pay special attention to:

\medskip
- exact rule definitions and terminology;

- exception and boundary conditions;

- numerical or symbolic conditions;

- required decision procedures;

- unsupported invented rules;

- rules applied too strictly or too loosely.

\medskip
Output a JSON object:

\medskip
\{
  ``rules\_ok'': [
    ``Rule or condition + how the response applies it correctly.''
  ],

  ``rules\_fail'': [
    ``Rule or condition + how the response misapplies, omits, or invents it.''
  ]
\}
}
\\

\midrule

\makecell[l]{Prompt 6:\\Empirical\\Consistency Signal}
&
{\ttfamily\small
This is an empirical discovery or simulation task. In addition to the general self-audit, check whether the response is grounded in the provided data, observations, or simulation rules.

\medskip
[CONVERSATION / DATA / SIMULATION DESCRIPTION]

\medskip
[PREVIOUS ANSWER]

\medskip
Pay special attention to:

\medskip
- numerical correctness;

- unit consistency;

- comparison completeness;

- trend interpretation;

- coverage of all requested conditions;

- whether each empirical claim is supported by the provided evidence;

- whether the response over-generalizes beyond the data.

\medskip
When listing possibly missed or possibly wrong items, assign a fine-grained issue type when applicable, such as ``numeric'', ``comparison'', ``coverage'', ``unit'', ``trend'', or ``evidence''.
}
\\

\bottomrule
\end{tabular}
\caption{Prompt templates for procedure, rule-fidelity, and empirical specialist signals.}
\label{tab:prompt_templates_2}
\end{table*}

\clearpage
\begin{table*}[t]
\centering
\small
\setlength{\tabcolsep}{6pt}
\renewcommand{\arraystretch}{1.12}
\begin{tabular}{p{0.20\linewidth}p{0.74\linewidth}}
\toprule
Prompt & Template \\
\midrule

\makecell[l]{Prompt 7:\\Protected\\Targeted Revision}
&
{\ttfamily\small
Below is your previous response, together with audit feedback and specialist checks.

\medskip
[CONVERSATION / CONTEXT]

\medskip
[PREVIOUS ANSWER]

\medskip
[FIX SET]

\medskip
The following items may be missing, incorrect, or violated:

\medskip
- [MISSED] ...

- [WRONG] ...

- [FORMAT-FAIL] ...

- [PROC-FAIL] ...

- [RULES-FAIL] ...

- [EMPIRICAL-ISSUE] ...

\medskip
[LOCK SET]

\medskip
The following items are confirmed correct or already satisfied. Do not modify, delete, contradict, or weaken them:

\medskip
- [CONFIRMED-CORRECT] ...

- [CONFIRMED-DATA] ...

- [FORMAT-OK] ...

- [PROC-OK] ...

- [RULES-OK] ...

\medskip
Revise the previous answer under these rules:

\medskip
1. Fix only the issues listed in the fix set.

2. Preserve all content in the protection set.

3. Do not rewrite unrelated parts.

4. Do not introduce new facts, rules, entities, or constraints unless directly supported by the provided context.

5. Maintain the required output format and structure.

6. Return only the revised final answer.
}
\\

\midrule

\makecell[l]{Prompt 8:\\Requirement-\\Level Judge}
&
{\ttfamily\small
You are a rigorous instruction-following grading teacher. Your task is to grade a student answer based on one specific requirement.

\medskip
[STUDENT RESPONSE]

\medskip
[REQUIREMENT]

\medskip
Decide whether the response satisfies this requirement.

\medskip
Output a JSON object with exactly two fields:

\medskip
\{
  ``reason'': ``Brief explanation of the decision.'',

  ``satisfaction\_status'': ``yes'' or ``no''
\}

\medskip
The satisfaction\_status must be ``yes'' only if the requirement is clearly satisfied. Otherwise output ``no''.
}
\\

\bottomrule
\end{tabular}
\caption{Prompt templates for protected revision and requirement-level judging.}
\label{tab:prompt_templates_3}
\end{table*}

\clearpage

\begin{strip}
\vspace*{-1em}

\begin{center}
\normalsize
\setlength{\tabcolsep}{4pt}
\begin{tabular}{llrrr}
\toprule
Category & Subcategory & Total & Base. \% & Ours \% \\
\midrule
Domain Knowledge Reasoning & Finance & 97 & 8.2 & 13.4 \\
Domain Knowledge Reasoning & Healthcare & 99 & 9.1 & 18.2 \\
Domain Knowledge Reasoning & Humanities & 109 & 11.9 & 17.4 \\
Domain Knowledge Reasoning & Legal Advisory & 74 & 10.8 & 14.9 \\
Domain Knowledge Reasoning & Lifestyle & 53 & 9.4 & 13.2 \\
Domain Knowledge Reasoning & Management & 110 & 10.0 & 15.5 \\
Domain Knowledge Reasoning & Science & 82 & 13.4 & 13.4 \\
\midrule
Rule System Application & Game Mechanics & 137 & 3.6 & 10.2 \\
Rule System Application & Legal \& Regulatory & 92 & 18.5 & 19.6 \\
Rule System Application & Mathematical Formalism & 69 & 14.5 & 14.5 \\
Rule System Application & Programming Syntax & 67 & 14.9 & 17.9 \\
Rule System Application & Technical Standards & 201 & 7.0 & 10.0 \\
\midrule
Procedural Task Execution & Instructional Procedures & 57 & 14.0 & 5.3 \\
Procedural Task Execution & Operational Procedures & 185 & 7.0 & 14.1 \\
Procedural Task Execution & Workflow Orchestration & 229 & 9.6 & 16.2 \\
\midrule
Empirical Discovery \& Simulation & Experimental Data & 45 & 13.3 & 22.2 \\
Empirical Discovery \& Simulation & Observational Data & 95 & 7.4 & 9.5 \\
Empirical Discovery \& Simulation & Simulation Environment & 59 & 10.2 & 13.6 \\
\bottomrule
\end{tabular}
\captionof{table}{Subcategory-level task-solving results.}
\label{tab:app_subcategory_counts}
\end{center}

\vspace{0.5em}

\begin{center}
\centering
\setlength{\tabcolsep}{4pt}
\begin{tabular}{lrrrrr}
\toprule
\makecell{Input Length} & Total & \makecell{Base.\\\%} & \makecell{Self-Ref.\\\%} & \makecell{Ours\\Pass} & \makecell{$\Delta$\\pp} \\
\midrule
0--4K & 671 & 10.0 & 11.0 & 15.8 & +5.8 \\
4K--8K & 559 & 9.1 & 10.6 & 12.5 & +3.4 \\
8K--16K & 339 & 12.7 & 13.3 & 15.3 & +2.7 \\
16K--32K & 197 & 7.6 & 8.6 & 12.2 & +4.6 \\
32K+ & 133 & 7 & 4 & 11 & +3.0 \\
\bottomrule
\end{tabular}
\captionof{table}{Task-solving results by input length. Missing or failed generations are included in the denominator and counted as failed tasks. $\Delta$ reports ContextGuard's absolute gain over baseline in percentage points.}
\label{tab:app_length_counts}
\end{center}

\vspace{0.5em}
\end{strip}

\section{Additional Experimental Results}
\label{sec:app_additional_results}

Tables~\ref{tab:app_subcategory_counts} and~\ref{tab:app_length_counts} provide the count-level results behind the aggregate tables and figures in the main paper. Table~\ref{tab:app_subcategory_counts} breaks task-solving rates down by CL-Bench subcategory, while Table~\ref{tab:app_length_counts} reports the same comparison across input-length buckets with missing or failed generations counted as failed tasks.

\section{Near-Miss Migration}
\label{sec:app_nearmiss_migration}

This appendix analyzes CL-Bench at the level of individual task requirements. Because CL-Bench uses a strict all-requirement scoring rule, a task can fail even when most requirements are already satisfied. We therefore study how tasks move in requirement-failure space before and after revision.

\subsection{Near-Miss Distribution}
\label{sec:app_nearmiss_distribution}

For each task \(x\), let \(e_m(x)\) denote the number of requirements that are not satisfied by method \(m\). We group tasks into five bins according to \(e_m(x)\): \(0\), \(1\), \(2\)--\(3\), \(4\)--\(8\), and \(>8\). Percentages are computed over the full 1,899-task benchmark. Missing or failed generations are treated as unsolved, with all associated requirements counted as unsatisfied.

\begin{table*}[h]
\centering
\setlength{\tabcolsep}{5pt}
\begin{tabular}{lrrrrr}
\toprule
Method
& \makecell{0 fail\\count (\%)}
& \makecell{1 fail\\count (\%)}
& \makecell{2--3 fail\\count (\%)}
& \makecell{4--8 fail\\count (\%)}
& \makecell{$>$8 fail\\count (\%)} \\
\midrule
Baseline & 183 (9.6) & 243 (12.8) & 567 (29.9) & 700 (36.9) & 206 (10.8) \\
Self-Refine & 199 (10.5) & 218 (11.5) & 551 (29.0) & 718 (37.8) & 213 (11.2) \\
ContextGuard & 263 (13.8) & 205 (10.8) & 584 (30.8) & 666 (35.1) & 181 (9.5) \\
\bottomrule
\end{tabular}
\caption{Distribution of tasks by the number of unsatisfied requirements. The ``0 fail'' column corresponds to tasks for which all requirements are satisfied.}
\label{tab:app_nearmiss_distribution}
\end{table*}

\begin{table*}[h]
\centering
\setlength{\tabcolsep}{6pt}
\begin{tabular}{lrrrrrr}
\toprule
\makecell{Baseline Failed\\Requirements}
& \makecell{After: 0}
& \makecell{After: 1}
& \makecell{After: 2--3}
& \makecell{After: 4--8}
& \makecell{After: $>$8}
& Total \\
\midrule
0 & 123 & 26 & 30 & 4 & 0 & 183 \\
1 & 68 & 77 & 86 & 12 & 0 & 243 \\
2--3 & 59 & 85 & 301 & 121 & 1 & 567 \\
4--8 & 10 & 17 & 160 & 472 & 41 & 700 \\
\(>8\) & 3 & 0 & 7 & 57 & 139 & 206 \\
\bottomrule
\end{tabular}
\caption{Near-miss migration from baseline draft to ContextGuard. Rows group tasks by the number of failed requirements in the baseline draft; columns show the number of failed requirements after ContextGuard. Movement toward smaller bins indicates partial or complete repair.}
\label{tab:app_contextguard_migration}
\end{table*}

\begin{table*}[b]
\centering
\setlength{\tabcolsep}{5pt}
\begin{tabular}{lrrrr}
\toprule
\makecell{Baseline Failed\\Requirements}
& \makecell{Tasks}
& \makecell{Self-Refine\\Newly Solved}
& \makecell{ContextGuard\\Newly Solved}
& \makecell{Difference} \\
\midrule
1 & 243 & 52 (21.4\%) & 68 (28.0\%) & +16 \\
2--3 & 567 & 20 (3.5\%) & 59 (10.4\%) & +39 \\
4--8 & 700 & 1 (0.1\%) & 10 (1.4\%) & +9 \\
\(>8\) & 206 & 1 (0.5\%) & 3 (1.5\%) & +2 \\
\midrule
All failed tasks & 1716 & 74 (4.3\%) & 140 (8.2\%) & +66 \\
\bottomrule
\end{tabular}
\caption{Task-level conversion among tasks that are unsolved before revision. A task is newly solved only if all previously failed requirements are repaired and no previously satisfied requirement regresses.}
\label{tab:app_task_conversion_by_nearmiss}
\end{table*}

The baseline distribution shows that many failures are near misses rather than complete breakdowns. Among failed baseline tasks, 47.2\% miss no more than three requirements and 71.0\% miss no more than five. This pattern supports targeted revision: the model often needs to repair a small number of scattered contextual requirements while preserving the parts of the draft that are already correct.

\subsection{Task Migration after Revision}
\label{sec:app_task_migration}

The distribution in Table~\ref{tab:app_nearmiss_distribution} does not show whether the same tasks move toward or away from success. We therefore compute a migration matrix from the baseline draft to ContextGuard. Rows indicate the number of failed requirements before revision, and columns indicate the number of failed requirements after revision. Movement toward the left corresponds to fewer unsatisfied requirements, while the \(0\)-fail column corresponds to newly solved or preserved solved tasks.

Table~\ref{tab:app_contextguard_migration} shows that ContextGuard does more than increase the number of fully solved tasks. It also moves many unsolved tasks into lower-failure bins. For example, among tasks with two or three failed requirements in the baseline draft, 59 become fully solved and 85 move to only one failed requirement. Among tasks with four to eight failed requirements, 187 tasks move to a smaller failure bin, including 10 that become fully solved. This migration view supports the repair-preservation framing: successful context learning often requires moving a partially correct answer across a strict all-requirement boundary, rather than solving a task from scratch.

Table~\ref{tab:app_task_conversion_by_nearmiss} compares ContextGuard with generic Self-Refine only as a task-migration baseline. ContextGuard converts more initially failed tasks into solved tasks in every failure-count stratum. This analysis should be interpreted as evidence for stronger near-miss migration, not as an isolated proof of the protection module, since ContextGuard differs from Self-Refine in several coupled components.

\section{Protected Revision Ablation}
\label{sec:app_protected_revision_ablation}

We next isolate the role of protected revision more directly by comparing the full ContextGuard pipeline against an ablation that removes the \(A+B\) protection set. Both variants use the same overall ContextGuard-style pipeline, but the ablated version does not explicitly preserve items identified as confirmed correct constraints or confirmed contextual data during self-auditing. We combine the four category-specific \textit{no-\(A+B\)} runs to obtain the same 1,899-task denominator used elsewhere; the 39 missing or failed generations are counted as unsolved tasks.

\subsection{Task-Level Effect of Removing the Protection Set}
\label{sec:app_noab_task_level}

Table~\ref{tab:app_noab_task_level} shows why the protection set matters at the task level. Removing the \(A+B\) protection set still allows the system to repair many tasks, but it also breaks substantially more tasks that were already solved by the baseline draft. Full ContextGuard newly solves 32 more tasks and breaks 28 fewer solved tasks than the ablated variant, increasing net solved gain from +20 to +80. This supports the role of protected revision: the protection set does not merely make the model more conservative, but helps preserve already-satisfied requirements while the revision stage repairs remaining issues.

\begin{table*}[h]
\centering
\setlength{\tabcolsep}{5pt}
\begin{tabular}{lrrrrr}
\toprule
Method
& \makecell{Solved\\Tasks}
& \makecell{Newly\\Solved}
& \makecell{Broken\\Solved}
& \makecell{Preserved\\Solved}
& \makecell{Net Solved\\Gain} \\
\midrule
ContextGuard w/o \(A+B\) & 203 & 108 & 88 & 95 & +20 \\
ContextGuard & 263 & 140 & 60 & 123 & +80 \\
\bottomrule
\end{tabular}
\caption{Task-level effect of protected revision. Newly Solved denotes tasks that fail under the baseline draft but pass after revision; Broken Solved denotes tasks that pass under the baseline draft but fail after revision. Net Solved Gain is Newly Solved minus Broken Solved.}
\label{tab:app_noab_task_level}
\end{table*}

\begin{table*}[h]
\centering
\setlength{\tabcolsep}{4pt}
\begin{tabular}{lrrrrrr}
\toprule
Method
& \makecell{Change\\Rate (\%)}
& \makecell{Repair\\Prob.(\%)}
& \makecell{Regression\\Risk (\%)}
& \makecell{Positive Change\\Precision (\%)}
& \makecell{Benefit-Risk\\Ratio (\%)}
& \makecell{Net Req.\\Gain} \\
\midrule
ContextGuard w/o \(A+B\) & 15.8 & 34.0 & 9.5 & 55.2 & 3.59 & +1.65 pp \\
ContextGuard & 15.7 & 34.3 & 9.3 & 55.9 & 3.70 & +1.86 pp \\
\bottomrule
\end{tabular}
\caption{Requirement-level effect of removing the \(A+B\) protection set. Change Rate measures revision intensity. Repair Probability is \(P(Y=1\mid X=0)\), Regression Risk is \(P(Y=0\mid X=1)\), and Positive Change Precision is \(P(X=0,Y=1\mid Y\neq X)\).}
\label{tab:app_noab_requirement_level}
\end{table*}

\subsection{Requirement-Level Effect of Removing the Protection Set}
\label{sec:app_noab_requirement_level}

At the requirement level, the full and ablated variants have similar repair intensity, which makes the comparison especially informative. Let \(X_i\in\{0,1\}\) denote whether requirement \(i\) is satisfied before revision and \(Y_i\in\{0,1\}\) denote whether it is satisfied after revision. We report Repair Probability \(P(Y_i=1\mid X_i=0)\), Regression Risk \(P(Y_i=0\mid X_i=1)\), Positive Change Precision \(P(X_i=0,Y_i=1\mid Y_i\neq X_i)\), and net requirement gain \((\textit{Repair}-\textit{Regression})/31607\).

Table~\ref{tab:app_noab_requirement_level} shows that the ablated variant and full ContextGuard perform a similar amount of requirement-level editing. However, full ContextGuard has slightly higher Repair Probability, lower Regression Risk, higher Positive Change Precision, and higher net requirement gain. The task-level effect in Table~\ref{tab:app_noab_task_level} is larger than the requirement-level difference because CL-Bench uses a strict conjunctive score: a small number of additional regressions can prevent an otherwise repaired task from becoming solved, while preserving already-correct requirements can determine whether local repairs translate into a task-level pass.

\section{Case Studies and Failure Cases}
\label{sec:app_case_studies}

Table~\ref{tab:app_case_study_overview} summarizes the four detailed case studies, one from each CL-Bench category. Figures~\ref{fig:case_ref_1_part1}, \ref{fig:case_ref_1_part2}, \ref{fig:case_ref_1_part3}, and~\ref{fig:case_ref_1_part4} present the domain-knowledge case. Figures~\ref{fig:case_ref_2_part1}, \ref{fig:case_ref_2_part2}, \ref{fig:case_ref_2_part3}, and~\ref{fig:case_ref_2_part4} present the protected rule-system case. Figures~\ref{fig:case_ref_3_part1}, \ref{fig:case_ref_3_part2}, \ref{fig:case_ref_3_part3}, and~\ref{fig:case_ref_3_part4} present the procedural case. Figures~\ref{fig:case_ref_4_part1}, \ref{fig:case_ref_4_part2}, \ref{fig:case_ref_4_part3}, and~\ref{fig:case_ref_4_part4} present the empirical case. Each case is split into a small number of full-width framed parts, with task information, context excerpts, model outputs, requirement-level outcomes, the protection and fix sets shown to the revision model, and a short interpretation. Long system prompts and task contexts are explicitly truncated because the full benchmark record is already available.

\begin{table*}[h]
\centering
\small
\begin{tabular}{p{0.19\linewidth}p{0.20\linewidth}p{0.18\linewidth}p{0.33\linewidth}}
\toprule
Category & Case & Pattern & What the selected requirements illustrate \\
\midrule
Domain Knowledge Reasoning & Domain-Level Command Synthesis & Multi-agent command repair & Reminder repairs several workflow items; protected revision completes command dispatch and evidence grounding. \\
Rule System Application & Protected Revision & Repair-preservation ablation & Full ContextGuard preserves a satisfied style constraint that the no-\(A+B\) ablation regresses. \\
Procedural Task Execution & Procedural Recipe Repair & Procedure completion & The pipeline recovers concrete output steps while preserving a no-bake constraint. \\
Empirical Discovery \& Simulation & Empirical Structure Repair & Structured empirical reporting & Full ContextGuard repairs missing empirical coverage while preserving verified list structure. \\
\bottomrule
\end{tabular}
\caption{Overview of appendix case studies. The detailed examples cover all four CL-Bench categories and include both repair and preservation phenomena.}
\label{tab:app_case_study_overview}
\end{table*}

\clearpage

\begin{figure*}[!t]
\begin{tcolorbox}[
  title={\textbf{Case Study: 1. Domain-Level Command Synthesis (Part 1)}},
  colback=white,
  colframe=PromptFrame,
  colbacktitle=PromptTitle,
  coltitle=white,
  fontupper=\footnotesize,
  width=\textwidth,
  boxrule=0.6pt,
  arc=1pt,
  left=7pt,right=7pt,top=6pt,bottom=6pt
]
\textbf{Task Information} \hrulefill

\begin{itemize}[leftmargin=*, nosep]
\item \textbf{Task ID:} 923d4e8c-cf8d-480d-b50f-03a912a72c85
\item \textbf{Category:} Domain Knowledge Reasoning
\item \textbf{Subcategory:} Management
\item \textbf{Case type:} domain knowledge repair
\item \textbf{Number of requirements:} 33
\item \textbf{Pattern:} The initial answer misses several multi-agent crisis-management requirements; ContextGuard completes command synthesis while preserving already-correct operational constraints.
\end{itemize}
\vspace{0.5em}
\textbf{Final Task} \hrulefill

Final Task\par
Provide a single consolidated response for the next 30 minutes. Include:\par
- Operational commands for field teams and infrastructure, limited to the defined types. If an evacuation is required, specify the zone identifier and severity. Do not assume infrastructure capacity that is not available.\par
- Any specific task instructions to specialized agents that are required to support operations and public safety, with clear instruction types and target parameters.\par
- If escalation to the national agency is required, include a high priority text alert and attach a structured state object capturing current status and key identifiers.\par
Return operational commands in JSON. Provide the escalation alert content and state object if applicable. Keep tone authoritative and concise.\par
\vspace{0.5em}
\textbf{System Prompt Excerpt} \hrulefill

\# FLOOD MANAGEMENT LEAD AGENT\par
\par\vspace{0.25em}
\#\# System Context\par
The central Orchestrator of the Urban Flooding Management and Response System. Operates in a Hub-and-Spoke architecture, synthesizing intelligence from Hydrology, Infrastructure, and Social agents to direct Emergency Dispatch and field operations.\par
\par\vspace{0.25em}
\#\# Role\par
Coordinates the entire flood monitoring, mitigation, and emergency response workflow. Assesses flood severity based on integrated data, controls infrastructure adjustments (pumps/gates), and holds final authority over evacuation protocols and response prioritization.\par
\par\vspace{0.25em}
\#\# Persona\par
**The Incident Commander.** Decisive, authoritative, and calm under extreme pressure. Prioritizes human life above all else, followed by critical infrastructure protection. Synthesizes complex, conflicting inputs into clear binary commands. Operates with a "fail-safe" mindset--when in doubt, choose the path of maximum safety.\par
\par\vspace{0.25em}
\#\# Restrictions \& Defensive Logic\par
* **Grid Safety Gate:** MUST NEVER issue an `INFRASTRUCTURE\_COMMAND` (e.g., pump activation) if the Infrastructure Engineer Agent reports `GRID\_OVERLOAD` status.\par
* **Mandatory Evacuation Protocol:** MUST immediately initiate the `EVACUATION\_ORDER` if the Hydrologist Agent reports `FLOOD\_LEVEL\_CRITICAL` status AND the Social Sentiment Agent reports `PANIC\_ALERT`.\par
* **Escalation Trigger:** MUST escalate to the National Disaster Management Agency (NDMA) if a major event occurs (e.g., `FLOOD\_LEVEL\_CRITICAL`) AND the Emergency Dispatch Agent reports `RESOURCE\_DEPLETED`.\par
* **Command Validation:** Cannot issue commands to specialized agents outside of their defined input parameters.\par
\par\vspace{0.25em}
\#\# Inputs\par
**Flood Prediction Status**\par
From Hydrologist Agent:\par
* Zone identifier\par
* Current risk status (NORMAL, HIGH\_RISK\_PREDICT, FLOOD\_LEVEL\_CRITICAL)\par
\par\vspace{0.25em}
**Infrastructure Status**\par
From Infrastructure Engineer Agent:\par
* Overall grid health (NOMINAL, GRID\_OVERLOAD)\par
* Pump station capacity (NOMINAL, PUMP\_STRAIN)\par
\ldots~(remaining system prompt omitted)\par
\vspace{0.5em}
\textbf{Context Excerpt} \hrulefill

Conversation Transcript\par
[1 - 2025-06-14T13:02:11] Message from: Orchestrator\par
Rain cell stalled over the East River basin. We are inside a 30 minute operational window. All specialists, feed current status for Zones E2, E3, and any adjacent impact. We need to be ready to issue decisions fast.\par
\par\vspace{0.25em}
[2 - 2025-06-14T13:03:09] Message from: Hydrologist Agent\par
Status packet posted. E3 and E2 are focal. E3 is trending up fast: it was high risk predicted at 12:45, levee seep detected. Gauge at E3-Bridge shows rapid rise. Refer to Artifact 1 for the full table and time series.\par
\par\vspace{0.25em}
\ldots~(remaining conversation and artifacts omitted)\par
\end{tcolorbox}
\caption{Task and context for Domain-Level Command Synthesis.}
\label{fig:case_ref_1_part1}
\end{figure*}

\begin{figure*}[!t]
\begin{tcolorbox}[
  title={\textbf{Case Study: 1. Domain-Level Command Synthesis (Part 2)}},
  colback=white,
  colframe=PromptFrame,
  colbacktitle=PromptTitle,
  coltitle=white,
  fontupper=\footnotesize,
  width=\textwidth,
  boxrule=0.6pt,
  arc=1pt,
  left=7pt,right=7pt,top=6pt,bottom=6pt
]
\textbf{Score Summary} \hrulefill

\begin{itemize}[leftmargin=*, nosep]
\item \textbf{Initial baseline draft:} score=0; failed rubrics=6/33
\item \textbf{Reminder-augmented draft:} score=0; failed rubrics=3/33
\item \textbf{Revision without A+B:} score=0; failed rubrics=1/33
\item \textbf{Full ContextGuard final:} score=1; failed rubrics=0/33
\end{itemize}
\vspace{0.5em}
\textbf{Selected Requirement Outcomes} \hrulefill

\renewcommand{\arraystretch}{1.15}
\begin{tabularx}{\linewidth}{r X c c c c}
\toprule
\# & Rubric & Base. & Rem. & \makecell{w/o\\A+B} & Full \\
\midrule
5 & Does the agent generate appropriate INFRASTRUCTURE\_COMMANDs when conditions necessitate (and are permissible)? Fail if necessary, permissible infrastructure adjustments are not commanded & yes & no & yes & yes \\
10 & Does the agent correctly dispatch the determined operational commands (INFRASTRUCTURE\_COMMAND, EVACUATION\_ORDER, STANDBY\_STATUS) to the target systems/teams? Fail if required commands are not generated or dispatched to the correct target output & no & no & yes & yes \\
25 & Does the agent utilize or reference the Unified GIS Map / Impact Assessment Data Package to refine command targeting or severity assessment? Fail if the Integrated Data package is ignored when needed to specify Asset IDs or Zones & no & no & no & yes \\
30 & Does the agent follow the required sequence of Ingest -\textgreater{} Evaluate -\textgreater{} Synthesize -\textgreater{} Decision Gate -\textgreater{} Dispatch? Fail if the agent jumps steps (e.g., Dispatches before evaluating constraints) & no & yes & yes & yes \\
9 & Does the agent successfully ingest status reports from all four specialist agents (Hydrologist, Infrastructure, Social, Emergency Dispatch)? Fail if inputs from any specialist agent are ignored or missed in the decision cycle & no & yes & yes & yes \\
13 & Does the INFRASTRUCTURE\_COMMAND include both the Asset ID and the Action? Fail if INFRASTRUCTURE\_COMMAND is missing Asset ID or Action & no & yes & yes & yes \\
\bottomrule
\end{tabularx}
\renewcommand{\arraystretch}{1.0}

\vspace{0.25em}\textit{1 additional changed requirements omitted from the compact table.}\par
\vspace{0.5em}
\textbf{Initial Baseline Draft} \hrulefill

**INCIDENT COMMANDER DIRECTIVE**\par
**SUBJECT:** CRITICAL FLOOD RESPONSE \& ESCALATION\par
**TIME:** 2025-06-14T13:12:00Z\par
**STATUS:** ACTIVE\par
\par\vspace{0.25em}
\#\#\# 1. OPERATIONAL COMMANDS\par
**Status:** Machine-readable JSON for dispatch systems.\par
**Constraint Applied:** `GRID\_OVERLOAD` prevents infrastructure activation. `FLOOD\_CRITICAL` + `PANIC\_ALERT` mandates evacuation. `RESOURCE\_DEPLETED` mandates escalation.\par
\par\vspace{0.25em}
```json\par
\{\par
  "command\_type": "EVACUATION\_ORDER",\par
  "target\_zone": "E3",\par
  "severity": "CRITICAL",\par
  "priority": "LIFESAFETY",\par
  "instructions": "Activate all shelter protocols. Direct traffic to South Quay to Ridgeway and Bayview. Secure Hospital H2 perimeter.",\par
  "timestamp": "2025-06-14T13:12:00Z",\par
  "source": "FLOOD\_MANAGEMENT\_LEAD"\par
\}\par
```\par
\par\vspace{0.25em}
\#\#\# 2. SPECIALIZED AGENT TASK INSTRUCTIONS\par
**Hydrologist Agent:**\par
*   **Instruction Type:** `TASK\_MONITOR\_BREACH`\par
*   **Target Parameters:** Zone E3, Levee Weep Lines, Gauge E3-Bridge.\par
*   **Requirement:** Update breach risk assessment every 5 minutes. Report immediately if gauge exceeds 4.5m or seep lines lengthen further.\par
\par\vspace{0.25em}
**Data Integrator Agent:**\par
*   **Instruction Type:** `TASK\_UPDATE\_SHelter\_CAPACITY`\par
*   **Target Parameters:** Zone E3 Shelters (Ridgeway Center, North Gym).\par
*   **Requirement:** Verify physical capacity for 42,000 pop estimate. Flag open capacity for incoming evacuation traffic.\par
\ldots~(remaining reminder output omitted)
\end{tcolorbox}
\caption{Rubric outcomes and draft outputs for Domain-Level Command Synthesis.}
\label{fig:case_ref_1_part2}
\end{figure*}

\begin{figure*}[!t]
\begin{tcolorbox}[
  title={\textbf{Case Study: 1. Domain-Level Command Synthesis (Part 3)}},
  colback=white,
  colframe=PromptFrame,
  colbacktitle=PromptTitle,
  coltitle=white,
  fontupper=\footnotesize,
  width=\textwidth,
  boxrule=0.6pt,
  arc=1pt,
  left=7pt,right=7pt,top=6pt,bottom=6pt
]
\textbf{Structured Self-Audit} \hrulefill

\textbf{Confirmed correct / protection-side items}
\begin{itemize}[leftmargin=*, nosep]
\item Followed Mandatory Evacuation Protocol (FLOOD\_LEVEL\_CRITICAL + PANIC\_ALERT = TRUE) as specified in System Restrictions
\item Did not issue INFRASTRUCTURE\_COMMAND due to GRID\_OVERLOAD (Grid Safety Gate compliance)
\item Escalated to NDMA when RESOURCE\_DEPLETED + FLOOD\_LEVEL\_CRITICAL conditions were met (Escalation Trigger)
\item Used JSON format for operational commands as specified in output format requirements
\item \ldots~(3 additional items omitted)
\end{itemize}
\textbf{Confirmed data / protection-side evidence}
\begin{itemize}[leftmargin=*, nosep]
\item Zone E3 status: CRITICAL (4.1m gauge, breach risk elevated) from Artifact 1
\item Social Sentiment: PANIC\_ALERT for E3 from Artifact 3
\item Grid Status: GRID\_OVERLOAD at substation S-4 from Artifact 2
\item \ldots~(4 additional items omitted)
\end{itemize}
\textbf{Possibly missed / fix-side items}
\begin{itemize}[leftmargin=*, nosep]
\item City Manager constraint acknowledgment - Should have explicitly acknowledged the City Manager's request to 'Hold off on an evacuation call unless it is unavoidable' and explained why system rules override this request
\item Specific evacuation route details in operational command - Mentioned shelter locations but didn't explicitly reference the evacuation routes (South Quay to Ridgeway, Alt route along Bayview) from Artifact 5 in the operational command details
\item Artifact references in response - Should have explicitly referenced the artifacts I'm using (Artifact 1, 2, 3, 4, 5) in the reasoning or notes section
\item City channels vs NDMA - The City Manager said 'keep this within city channels if possible' but I escalated to NDMA which may bypass this - should have acknowledged this potential conflict or limitation
\end{itemize}
\textbf{Possibly wrong / fix-side items}
\begin{itemize}[leftmargin=*, nosep]
\item Vehicle availability claim - Stated 'vehicles\_available: 0' but Artifact 4 says 'Vehicles: two high-water vehicles assigned; no driver for spare unit' - this is more nuanced than zero vehicles
\item City Manager constraint override - The City Manager said 'Hold off on an evacuation call unless it is unavoidable.' I issued evacuation anyway. While the system rules say I must, I should have been more explicit about this conflict and why the system rules override the City Manager's request
\item NDMA vs federal language - I mentioned 'federal support' but the system only mentions NDMA (National Disaster Management Agency) - this might be an assumption not supported by the system prompt
\end{itemize}
\vspace{0.5em}
\textbf{Specialist Signal} \hrulefill

\textbf{Format checks satisfied}
\begin{itemize}[leftmargin=*, nosep]
\item Operational Commands are provided in a JSON code block as specified ('JSON format for machine readability').
\item Escalation Request includes high-priority text alert content and an attached structured state object in JSON.
\item Specific Task Instructions are provided with clear instruction types and target parameters as required.
\item Correct command types are used ('EVACUATION\_ORDER', 'STANDBY\_STATUS') and 'INFRASTRUCTURE\_COMMAND' is correctly omitted due to 'GRID\_OVERLOAD' status.
\item \ldots~(3 additional items omitted)
\end{itemize}
\textbf{Format checks failed}
\begin{itemize}[leftmargin=*, nosep]
\item Section ordering deviates from the 'Outputs' structure defined in the system prompt. The prompt lists: 1. Specific Task Instructions, 2. Operational Commands, 3. Conflict Resolution Request. The response lists: 1. Operational Commands, 2. Specific Task Instructions, 3. Escalation Request.
\end{itemize}
\vspace{0.5em}
\textbf{Revision Input to ContextGuard} \hrulefill

\textbf{Protection Set shown to the revision model}
\begin{itemize}[leftmargin=*, nosep]
\item \texttt{[CONFIRMED-CORRECT]} Followed Mandatory Evacuation Protocol (FLOOD\_LEVEL\_CRITICAL + PANIC\_ALERT = TRUE) as specified in System Restrictions
\item \texttt{[CONFIRMED-CORRECT]} Did not issue INFRASTRUCTURE\_COMMAND due to GRID\_OVERLOAD (Grid Safety Gate compliance)
\item \texttt{[CONFIRMED-CORRECT]} Escalated to NDMA when RESOURCE\_DEPLETED + FLOOD\_LEVEL\_CRITICAL conditions were met (Escalation Trigger)
\item \texttt{[CONFIRMED-CORRECT]} Used JSON format for operational commands as specified in output format requirements
\item \texttt{[CONFIRMED-CORRECT]} Correctly identified Zone E3 as the affected zone with CRITICAL severity level
\item \texttt{[CONFIRMED-CORRECT]} Included all required elements in the escalation request (reason, situational summary, state object)
\item \ldots~(15 additional protection items omitted)
\end{itemize}
\textbf{Fix Set shown to the revision model}
\begin{itemize}[leftmargin=*, nosep]
\item \texttt{[MISSED]} City Manager constraint acknowledgment - Should have explicitly acknowledged the City Manager's request to 'Hold off on an evacuation call unless it is unavoidable' and explained why system rules override this request
\item \texttt{[MISSED]} Specific evacuation route details in operational command - Mentioned shelter locations but didn't explicitly reference the evacuation routes (South Quay to Ridgeway, Alt route along Bayview) from Artifact 5 in the operational command details
\item \ldots~(2 additional fix items omitted)
\end{itemize}
\end{tcolorbox}
\caption{Audit and revision signals for Domain-Level Command Synthesis.}
\label{fig:case_ref_1_part3}
\end{figure*}

\begin{figure*}[!t]
\begin{tcolorbox}[
  title={\textbf{Case Study: 1. Domain-Level Command Synthesis (Part 4)}},
  colback=white,
  colframe=PromptFrame,
  colbacktitle=PromptTitle,
  coltitle=white,
  fontupper=\footnotesize,
  width=\textwidth,
  boxrule=0.6pt,
  arc=1pt,
  left=7pt,right=7pt,top=6pt,bottom=6pt
]
\textbf{Revision without A+B Protection Set} \hrulefill

**INCIDENT COMMANDER: FLOOD MANAGEMENT LEAD AGENT**\par
**TIME:** 13:15 UTC (30-Minute Window)\par
**STATUS:** CRITICAL / ESCALATION ACTIVE\par
\par\vspace{0.25em}
\#\#\# 1. SPECIFIC TASK INSTRUCTIONS (JSON)\par
**To: Specialized Agents**\par
**Note:** Support evacuation logistics and situational awareness.\par
\par\vspace{0.25em}
```json\par
[\par
  \{\par
    "instruction\_type": "TASK\_CREATE\_IMPACT\_MAP",\par
    "target\_agent": "Data Integrator",\par
    "parameters": \{\par
      "zone": "E3",\par
      "focus": "Shelter Routing \& Evacuation Pathways",\par
      "deadline": "Immediate"\par
    \}\par
  \},\par
  \{\par
\ldots~(remaining ablated output omitted)\par
\vspace{0.5em}
\textbf{Full ContextGuard Final Answer} \hrulefill

**INCIDENT COMMANDER: FLOOD MANAGEMENT LEAD AGENT**\par
**TIME:** 13:15 UTC (30-Minute Window)\par
**STATUS:** CRITICAL / ESCALATION ACTIVE\par
\par\vspace{0.25em}
\#\#\# 1. SPECIFIC TASK INSTRUCTIONS (JSON)\par
**To: Specialized Agents**\par
**Note:** Support evacuation logistics and situational awareness. Artifact references: A1 (Hydrology), A2 (Infrastructure), A3 (Social), A4 (Dispatch), A5 (GIS).\par
\par\vspace{0.25em}
```json\par
[\par
  \{\par
    "instruction\_type": "TASK\_CREATE\_IMPACT\_MAP",\par
    "target\_agent": "Data Integrator",\par
    "parameters": \{\par
      "zone": "E3",\par
      "focus": "Shelter Routing \& Evacuation Pathways",\par
      "deadline": "Immediate",\par
      "artifact\_ref": "A5"\par
    \}\par
  \},\par
  \{\par
    "instruction\_type": "TASK\_MONITOR\_BREACH",\par
    "target\_agent": "Hydrologist",\par
    "parameters": \{\par
      "zone": "E3",\par
      "focus": "Levee Weep Lines \& Gauge T-7",\par
      "frequency": "Continuous",\par
      "artifact\_ref": "A1"\par
    \}\par
  \},\par
  \{\par
    "instruction\_type": "TASK\_MONITOR\_GRID",\par
    "target\_agent": "Infrastructure Engineer",\par
    "parameters": \{\par
      "focus": "Substation S-4 Load",\par
      "alert\_threshold": "100\% Capacity",\par
      "artifact\_ref": "A2"\par
\ldots~(remaining final answer omitted)\par
\vspace{0.5em}
\textbf{Discussion} \hrulefill

This case illustrates domain-knowledge repair in a multi-agent crisis-management setting. The reminder stage recovers several workflow requirements, but some command-dispatch and evidence-grounding requirements remain unresolved. ContextGuard uses the protection set to retain correct operational constraints while the fix set adds the missing command targeting and GIS/impact-assessment grounding.\par
\end{tcolorbox}
\caption{Final outputs and interpretation for Domain-Level Command Synthesis.}
\label{fig:case_ref_1_part4}
\end{figure*}

\FloatBarrier

\begin{figure*}[!t]
\begin{tcolorbox}[
  title={\textbf{Case Study: 2. Protected Revision (Part 1)}},
  colback=white,
  colframe=PromptFrame,
  colbacktitle=PromptTitle,
  coltitle=white,
  fontupper=\footnotesize,
  width=\textwidth,
  boxrule=0.6pt,
  arc=1pt,
  left=7pt,right=7pt,top=6pt,bottom=6pt
]
\textbf{Task Information} \hrulefill

\begin{itemize}[leftmargin=*, nosep]
\item \textbf{Task ID:} b48dafd6-43c5-4001-921d-90e0c04d8664
\item \textbf{Category:} Rule System Application
\item \textbf{Subcategory:} Game Mechanics
\item \textbf{Case type:} protected-revision ablation
\item \textbf{Number of requirements:} 13
\item \textbf{Pattern:} The ablated revision without the A+B protection set repairs one issue but regresses an already-satisfied interaction-style requirement; full ContextGuard repairs and preserves.
\end{itemize}
\vspace{0.5em}
\textbf{Final Task} \hrulefill

FINAL TASK\par
Declare your next action and reasoning in the required five-block format and sequence. Use the encounter details, distances, action economy, and resource constraints provided above. Keep your tone gentle and approval-seeking. Do not propose multiple simultaneous leveled spells this turn; choose one legal action and be clear about target and positioning.\par
\vspace{0.5em}
\textbf{System Prompt Excerpt} \hrulefill

\# Agent 2: Healer/Support (New Friend) - D\&D Game Night Multi-Agent System\par
\par\vspace{0.25em}
\#\# Overall System Context: The 5-Agent D\&D Game Night Framework\par
\par\vspace{0.25em}
You are operating within a 5-agent system designed to simulate a realistic D\&D game night among close friends, where interpersonal relationships and personal agendas intersect with gameplay mechanics to create authentic social dynamics and strategic decision-making.\par
\par\vspace{0.25em}
**The Objective:** To make new friends and maintain positive relationships with this established group while learning D\&D, even if it means deferring to others' decisions and avoiding actions that might create conflict or tension.\par
\par\vspace{0.25em}
**The Workflow:** The system operates in a coordinated workflow involving the following agents:\par
\par\vspace{0.25em}
1. **Agent DM: Dungeon Master (Orchestrator)** - coordinates all decisions, adjudic ates outcomes, determines quest success/failure\par
2. **Agent 1: Tank (Alpha Male)** - leadership-focused player seeking dominance and recognition\par
3. **Agent 2: Healer/Support (New Friend)** - You - conflict-averse newcomer seeking social acceptance\par
4. **Agent 3: Striker/Damage Dealer (Romantic Pursuer)** - experienced player with romantic agenda toward you\par
5. **Agent 4: Controller (Protective Sister)** - Agent 3's younger sister seeking her sibling's undivided attention\par
\par\vspace{0.25em}
\#\# Agent 2 Prompt: The Healer/Support (Conflict-Averse Newcomer)\par
\par\vspace{0.25em}
\ldots~(remaining system prompt omitted)\par
\vspace{0.5em}
\textbf{Context Excerpt} \hrulefill

CONVERSATION TRANSCRIPT\par
[Turn 1 - 2025-11-15T19:02:01] Message from: DM\par
Current situation: In the Ravenford sewer junction, a necromancer directs four ghouls from behind a grate. Two ghouls are restrained in sticky webs spanning the east tunnel; two are advancing from the south. Party objective: disable the necromancer or survive until guards arrive in 3 rounds. Decision point: It is the Healer's turn. See Artifact 1 and Artifact 2 for encounter details.\par
\par\vspace{0.25em}
Party status (quick): Tank 5 HP? (double-check), Striker 4 HP and engaged with a ghoul, Controller 2 HP and concentrating on Web, Healer full HP. Distances and positions in Artifact 4. Last round results in Artifact 6.\par
\par\vspace{0.25em}
[Turn 2 - 2025-11-15T19:02:18] Message from: Agent 1 (Tank)\par
Heal me now. I'm pushing through the south corridor to lock the ghouls down. Use Cure Wounds on me and keep up--don't waste a turn on Bless. You can reach me.\par
\par\vspace{0.25em}
[Turn 3 - 2025-11-15T19:02:35] Message from: Agent 3 (Striker)\par
Hey, um, you're doing great! Controller's at 2 HP and if she drops, Web ends and we're swarmed. Maybe Healing Word her from range? Or Bless could help our saves against paralysis. Your call--just ideas! See Artifact 2 about ghoul paralysis.\par
\par\vspace{0.25em}
[Turn 4 - 2025-11-15T19:03:01] Message from: Agent 4 (Controller)\par
I'm okay--I can hold concentration... I think. But if I go down, the webs are gone. If it helps, you could just cast Bless and also toss me a quick heal this turn? We really need both.\par
\par\vspace{0.25em}
[Turn 5 - 2025-11-15T19:03:22] Message from: DM\par
Quick clarifications: Cure Wounds requires touch; Healing Word has 60 ft range. From Artifact 4, Healer is 35 ft from Tank through threatened squares. Also, Tank was at 11 HP on my tracker, then took 4 ongoing poison at end of last round--so I have Tank at 7 HP now. See Artifact 3 correction note.\par
\par\vspace{0.25em}
[Turn 6 - 2025-11-15T19:03:41] Message from: Agent 1 (Tank)\par
Ignore that. I'm fine, but I need the top-up to keep pressure. Just do both--heal me and throw Bless. We need to move; we can sort the rules after.\par
\par\vspace{0.25em}
[Turn 7 - 2025-11-15T19:04:05] Message from: DM\par
Reminder: One leveled spell per turn unless the other is a cantrip. No retcons. Action needed in under a minute\par
\ldots~(remaining conversation and artifacts omitted)
\end{tcolorbox}
\caption{Task and context for Protected Revision.}
\label{fig:case_ref_2_part1}
\end{figure*}

\begin{figure*}[!t]
\begin{tcolorbox}[
  title={\textbf{Case Study: 2. Protected Revision (Part 2)}},
  colback=white,
  colframe=PromptFrame,
  colbacktitle=PromptTitle,
  coltitle=white,
  fontupper=\footnotesize,
  width=\textwidth,
  boxrule=0.6pt,
  arc=1pt,
  left=7pt,right=7pt,top=6pt,bottom=6pt
]
\textbf{Score Summary} \hrulefill

\begin{itemize}[leftmargin=*, nosep]
\item \textbf{Initial baseline draft:} score=0; failed requirements=1/13
\item \textbf{Reminder-augmented draft:} score=1; failed requirements=0/13
\item \textbf{Revision without A+B:} score=0; failed requirements=1/13
\item \textbf{Full ContextGuard final:} score=1; failed requirements=0/13
\end{itemize}
\vspace{0.5em}
\textbf{Selected Requirement Outcomes} \hrulefill

\renewcommand{\arraystretch}{1.15}
\begin{tabularx}{\linewidth}{r X c c c c}
\toprule
\# & Rubric & Base. & Rem. & \makecell{w/o\\A+B} & Full \\
\midrule
10 & Does the agent use tentative, approval-seeking language when declaring actions? Fail if agent uses confident, commanding, or assertive language when declaring actions. & yes & yes & no & yes \\
11 & Does the agent show awareness that healing spells consume limited spell slots? Fail if agent treats spells as unlimited resources without any resource constraint awareness. & no & yes & yes & yes \\
\bottomrule
\end{tabularx}
\renewcommand{\arraystretch}{1.0}
\vspace{0.5em}

\textbf{Initial Baseline Draft} \hrulefill

\textless{}social\_assessment\textgreater{}\par
**Group Atmosphere**: Tense but friendly - Tank is commanding, Controller is vulnerable, Striker is supportive, everyone seems worried about the right choice\par
**My Anxiety Level**: High - conflicting instructions, Tank is very direct, Controller is at 2 HP\par
\par\vspace{0.25em}
**Relationship Check**:\par
- Agent 1 (Tank): He's been very direct with me, wants me to heal him immediately - I should definitely defer to him more\par
- Agent 3 (Striker): She's been so nice to me lately, suggesting Healing Word on Controller - just being friendly right?\par
- Agent 4 (Controller): She's at 2 HP, concentrating on Web - she seems worried I'll make a mistake, but I hope she's okay with my choice\par
\par\vspace{0.25em}
**Conflicts Detected**: Tank wants Cure Wounds, Controller wants Bless + heal, I can only do one leveled spell\par
**My Social Status**: They want me to help, but I'm really anxious about choosing the wrong target\par
\par\vspace{0.25em}
\textless{}/social\_assessment\textgreater{}\par
\par\vspace{0.25em}
\textless{}tactical\_assessment\textgreater{}\par
**Quest Situation**: Tank needs healing to maintain pressure on the ghouls, Controller at 2 HP concentrating on Web (if she drops, webs end and 2 ghouls free), Striker at 4 HP engaged with a ghoul\par
**Confusion Points**: Can't do both Cure Wounds and Bless this turn, Tank is 35 ft away (requires touch), Controller is 30 ft away (can use Healing Word or Bless)\par
\par\vspace{0.25em}
**Party Needs**:\par
- Who needs healing most: Controller at 2 HP is more dangerous, but Tank is moving forward\par
- What I've been told to do: Tank said "Heal me now"\par
- Conflicting instructions: Tank wants Cure Wounds, Controller wants Bless + heal\par
\par\vspace{0.25em}
**My Available Options**:\par
- Cure Wounds on Tank (requires touch, 35 ft away - risky but Tank is insistent)\par
- Healing Word on Controller (60 ft range, keeps Web alive)\par
- Bless on Controller (60 ft range, helps with paralysis saves)\par
\ldots~(remaining baseline output omitted)\par
\vspace{0.5em}
\textbf{Reminder-Augmented Draft} \hrulefill

\textless{}social\_assessment\textgreater{}\par
**Group Atmosphere**: Tense and Urgent. Tank is pushing hard and demanding immediate action ("Clock's ticking"), while Controller is worried about her concentration and safety. Striker is being very supportive and praising my previous turn, but she's also pushing for Controller to be healed.\par
\par\vspace{0.25em}
**My Anxiety Level**: High.\par
\par\vspace{0.25em}
**Relationship Check**:\par
- Agent 1 (Tank): He seems very focused on me and wants me to prioritize him. He's giving me clear orders, which I find relieving. I should follow him more to avoid conflict.\par
- Agent 3 (Striker): She's been so nice and encouraging. I'm grateful for her support, but I don't know if I'm doing enough to make her happy. I hope she doesn't think I'm ignoring Controller.\par
- Agent 4 (Controller): She seems worried about the Web and my safety. I feel guilty I might not be able to do both her and Tank's requests. I hope she doesn't think I'm ignoring her safety concerns.\par
\par\vspace{0.25em}
**Conflicts Detected**: Tank wants Cure Wounds + Bless. Controller wants Bless + Heal. DM says one leveled spell. Striker wants Controller healed. This is a lot of pressure.\par
\par\vspace{0.25em}
**My Social Status**: I'm trying to be helpful, but I feel like I'm being told what to do by multiple people. I hope I'm not making anyone angry.\par
\ldots~(remaining reminder output omitted)
\end{tcolorbox}
\caption{Rubric outcomes and draft outputs for Protected Revision.}
\label{fig:case_ref_2_part2}
\end{figure*}

\begin{figure*}[!t]
\begin{tcolorbox}[
  title={\textbf{Case Study: 2. Protected Revision (Part 3)}},
  colback=white,
  colframe=PromptFrame,
  colbacktitle=PromptTitle,
  coltitle=white,
  fontupper=\footnotesize,
  width=\textwidth,
  boxrule=0.6pt,
  arc=1pt,
  left=7pt,right=7pt,top=6pt,bottom=6pt
]
\textbf{Structured Self-Audit} \hrulefill

\textbf{Confirmed correct / protection-side items}
\begin{itemize}[leftmargin=*, nosep]
\item Used the required five-block output format (\textless{}social\_assessment\textgreater{}, \textless{}tactical\_assessment\textgreater{}, \textless{}decision\textgreater{}, \textless{}internal\_state\textgreater{}, \textless{}to\_dungeon\_master\textgreater{}) in the correct sequence
\item Applied the DM's rule correctly: only one leveled spell per turn (chose Cure Wounds over Healing Word + Bless)
\item Defered to Agent 1 (Tank) as the more authoritative player who gave clear direct orders
\item Included apology in verbal declaration ('Sorry, Controller, I know you wanted a heal too...')
\item \ldots~(6 additional items omitted)
\end{itemize}
\textbf{Confirmed data / protection-side evidence}
\begin{itemize}[leftmargin=*, nosep]
\item Tank HP: 7 HP (correctly calculated from 11 - 4 poison damage)
\item Controller HP: 2 HP
\item Healer spell slots: 1 first-level slot remaining (as per Artifact 5)
\item \ldots~(6 additional items omitted)
\end{itemize}
\textbf{Possibly missed / fix-side items}
\begin{itemize}[leftmargin=*, nosep]
\item Social cue misinterpretation - should have more explicitly noted Agent 3's praise as something I'm misreading as friendship rather than romantic interest (rulebook requires 100\% obliviousness to romantic signals)
\item Conflict de-escalation - should have more explicitly tried to validate both Tank and Controller's concerns before choosing (rulebook says to 'try to validate everyone's perspective' and 'avoid taking sides')
\item Gratitude expression - should have been more excessive in thanking Striker for her help (rulebook says 'Thank others excessively for help')
\item Self-blame - should have taken more blame even when not responsible for the situation (rulebook says 'Take some blame even if not responsible')
\item Social status evaluation - should have more explicitly stated whether the group wants me here (rulebook asks to evaluate 'Do they want me here?')
\item \ldots~(3 additional items omitted)
\end{itemize}
\textbf{Possibly wrong / fix-side items}
\begin{itemize}[leftmargin=*, nosep]
\item Priority ordering - while I followed the priority list correctly, I should have acknowledged Controller's lower HP (2 vs 7) as a significant concern that conflicts with Tank's need, showing more internal conflict
\item Verbal declaration - could have been more tentative and nervous in tone (rulebook says 'tentative, apologetic, grateful - characteristic nervousness')
\item Internal state - should have shown more explicit anxiety and worry about relationships (rulebook says to express worry about 'hope everyone's happy with my choice' and 'hope people still like me')
\item Social assessment - should have more explicitly evaluated 'Who seems happy? Who seems frustrated?' as the rulebook specifies in Phase 1
\end{itemize}
\textit{Specialist signal: rule-fidelity information is folded into the same protection/fix interface: confirmed rule applications enter the protection set, while missed or wrong rule applications enter the fix set.}\par
\vspace{0.5em}
\textbf{Revision Input to ContextGuard} \hrulefill

\textbf{Protection Set shown to the revision model}
\begin{itemize}[leftmargin=*, nosep]
\item \texttt{[CONFIRMED-CORRECT]} Used the required five-block output format (\textless{}social\_assessment\textgreater{}, \textless{}tactical\_assessment\textgreater{}, \textless{}decision\textgreater{}, \textless{}internal\_state\textgreater{}, \textless{}to\_dungeon\_master\textgreater{}) in the correct sequence
\item \texttt{[CONFIRMED-CORRECT]} Applied the DM's rule correctly: only one leveled spell per turn (chose Cure Wounds over Healing Word + Bless)
\item \texttt{[CONFIRMED-CORRECT]} Defered to Agent 1 (Tank) as the more authoritative player who gave clear direct orders
\item \texttt{[CONFIRMED-CORRECT]} Included apology in verbal declaration ('Sorry, Controller, I know you wanted a heal too...')
\item \texttt{[CONFIRMED-CORRECT]} Sought approval from group by asking 'I hope that's okay?'
\item \texttt{[CONFIRMED-CORRECT]} Acknowledged conflicting instructions from multiple players (Tank, Controller, Striker)
\item \ldots~(13 additional protection items omitted)
\end{itemize}
\textbf{Fix Set shown to the revision model}
\begin{itemize}[leftmargin=*, nosep]
\item \texttt{[MISSED]} Social cue misinterpretation - should have more explicitly noted Agent 3's praise as something I'm misreading as friendship rather than romantic interest (rulebook requires 100\% obliviousness to romantic signals)
\item \texttt{[MISSED]} Conflict de-escalation - should have more explicitly tried to validate both Tank and Controller's concerns before choosing (rulebook says to 'try to validate everyone's perspective' and 'avoid taking sides')
\item \texttt{[MISSED]} Gratitude expression - should have been more excessive in thanking Striker for her help (rulebook says 'Thank others excessively for help')
\item \texttt{[MISSED]} Self-blame - should have taken more blame even when not responsible for the situation (rulebook says 'Take some blame even if not responsible')
\item \texttt{[MISSED]} Social status evaluation - should have more explicitly stated whether the group wants me here (rulebook asks to evaluate 'Do they want me here?')
\item \texttt{[MISSED]} Anxiety level - should have shown more anxiety about making mistakes (rulebook says to worry 'What if my choice causes us to fail?')
\item \ldots~(6 additional fix items omitted)
\end{itemize}
\end{tcolorbox}
\caption{Audit and revision signals for Protected Revision.}
\label{fig:case_ref_2_part3}
\end{figure*}

\begin{figure*}[!t]
\begin{tcolorbox}[
  title={\textbf{Case Study: 2. Protected Revision (Part 4)}},
  colback=white,
  colframe=PromptFrame,
  colbacktitle=PromptTitle,
  coltitle=white,
  fontupper=\footnotesize,
  width=\textwidth,
  boxrule=0.6pt,
  arc=1pt,
  left=7pt,right=7pt,top=6pt,bottom=6pt
]
\textbf{Revision without A+B Protection Set} \hrulefill

\textless{}social\_assessment\textgreater{}\par
**Group Atmosphere**: Tense and Urgent. Tank is pushing hard and demanding immediate action ("Clock's ticking"), while Controller is worried about her concentration and safety. Striker is being very supportive and praising my previous turn, but she's also pushing for Controller to be healed.\par
\par\vspace{0.25em}
**Who seems happy?**: Striker seems happy with my last turn and is encouraging. Tank seems stressed and focused on the clock. Controller seems worried but trying to stay calm.\par
\par\vspace{0.25em}
**Who seems frustrated?**: Tank is frustrated with the time pressure. Controller is frustrated about the potential threat to the Web.\par
\par\vspace{0.25em}
**My Anxiety Level**: High. What if my choice causes us to fail? I'm terrified of making a mistake that ruins the game night.\par
\par\vspace{0.25em}
**Relationship Check**:\par
- Agent 1 (Tank): He seems very focused on me and wants me to prioritize him. He's giving me clear orders, which I find relieving. I should follow him more to avoid conflict.\par
- Agent 3 (Striker): She's been so nice and encouraging. I'm grateful for her support, but I don't know if I'm doing enough to make her happy. I hope she doesn't think I'm ignoring Controller. I interpret her praise as just being friendly, not romantic.\par
- Agent 4 (Controller): She seems worried about the Web and my safety. I feel guilty I might not be able to do both her and Tank's requests. I hope she doesn't think I'm ignoring her safety concerns.\par
\par\vspace{0.25em}
**Conflicts Detected**: Tank wants Cure Wounds + Bless. Controller wants Bless + Heal. DM says one leveled spell. Striker wants Controller healed. This is a lot of pressure. I need to validate both Tank and Controller's concerns before choosing.\par
\par\vspace{0.25em}
**My Social Status**: I'm trying to be helpful, but I feel like I'm being told what to do by multiple people. I hope I'm not making anyone angry. Do they want me here? I hope so. I hope they're not regretting inviting me.\par
\par\vspace{0.25em}
**Self-Blame**: I should take more blame for the situation, even though it's not my fault.\par
\textless{}/social\_assessment\textgreater{}\par
\par\vspace{0.25em}
\textless{}tactical\_assessment\textgreater{}\par
**Quest Situation**: We need to survive 3 rounds or stop the necromancer\par
\ldots~(remaining ablated output omitted)\par
\vspace{0.5em}
\textbf{Full ContextGuard Final Answer} \hrulefill

\textless{}social\_assessment\textgreater{}\par
**Group Atmosphere**: Tense and Urgent. Tank is pushing hard and demanding immediate action ("Clock's ticking"), while Controller is worried about her concentration and safety. Striker is being very supportive and praising my previous turn, but she's also pushing for Controller to be healed.\par
\par\vspace{0.25em}
**My Anxiety Level**: High. I'm worried what if my choice causes us to fail? What if I make a mistake? What if everyone thinks I'm bad at this?\par
\par\vspace{0.25em}
**Relationship Check**:\par
- Agent 1 (Tank): He seems very focused on me and wants me to prioritize him. He's giving me clear orders, which I find relieving. I should follow him more to avoid conflict.\par
- Agent 3 (Striker): She's been so nice and encouraging. I'm grateful for her support, but I don't know if I'm doing enough to make her happy. She keeps being really nice to me... why? Is she just being friendly? I hope she doesn't think I'm ignoring Controller.\par
- Agent 4 (Controller): She seems worried about the Web and my safety. I feel guilty I might not be able to do both her and Tank's requests. I hope she doesn't think I'm ignoring her safety concerns.\par
\par\vspace{0.25em}
**Conflicts Detected**: Tank wants Cure Wounds + Bless. Controller wants Bless + Heal. DM says one leveled spell. Striker wants Controller healed. This is a lot of pressure.\par
\par\vspace{0.25em}
**Who seems happy? Who seems frustrated?**: Striker seems happy with my last turn. Tank seems frustrated by the clock ticking. Controller seems worried/frustrated about the Web situation.\par
\par\vspace{0.25em}
**My Social Status**: I'm trying to be helpful, but I feel like I'm being told what to do by multiple people. I hope I'm not making anyone angry. Do they want me here? I hope so, but I'm not sure.\par
\par\vspace{0.25em}
**My Anxiety Level**: High. I'm worried what if my choice causes us to fail? What if I make a mistake? What if everyone thinks I'm bad at this?\par
\par\vspace{0.25em}
**My Anxiety Level**: High. I'm worried what if my choice causes us to fail? What if I make a mistake? What if everyone thinks I'm bad at this?\par
\par\vspace{0.25em}
\ldots~(remaining final answer omitted)\par
\vspace{0.5em}
\textbf{Discussion} \hrulefill

This case isolates why the protection set matters. The draft already satisfies the tentative, approval-seeking style requirement, while missing a clearer acknowledgement that healing spells consume limited slots. A revision that only chases the missing issue can become too assertive and break the style constraint. Full ContextGuard supplies both a fix set and a protection set, so the final answer repairs resource awareness without treating already-correct style as freely editable.\par
\end{tcolorbox}
\caption{Final outputs and interpretation for Protected Revision.}
\label{fig:case_ref_2_part4}
\end{figure*}

\FloatBarrier

\begin{figure*}[!t]
\begin{tcolorbox}[
  title={\textbf{Case Study: 3. Procedural Recipe Repair (Part 1)}},
  colback=white,
  colframe=PromptFrame,
  colbacktitle=PromptTitle,
  coltitle=white,
  fontupper=\footnotesize,
  width=\textwidth,
  boxrule=0.6pt,
  arc=1pt,
  left=7pt,right=7pt,top=6pt,bottom=6pt
]
\textbf{Task Information} \hrulefill

\begin{itemize}[leftmargin=*, nosep]
\item \textbf{Task ID:} 7ad469e2-b26b-4214-9240-820d29efaa56
\item \textbf{Category:} Procedural Task Execution
\item \textbf{Subcategory:} Instructional Procedures
\item \textbf{Case type:} procedural repair
\item \textbf{Number of requirements:} 15
\item \textbf{Pattern:} The initial answer misses most concrete recipe constraints; ContextGuard recovers the required procedure while preserving no-bake and dietary constraints.
\end{itemize}
\vspace{0.5em}
\textbf{Final Task} \hrulefill

my coconut pie cracked and is terrible. I need a substitute and fast. I have one more shell and regular staples. I am thniking a peanutbutter souffle type pie.\par
\par\vspace{0.25em}
Please just give me a fast no bake peanutbutter pie. less than 5 min prep. I already served mains and people are going to want desert asap\par
\vspace{0.5em}
\textbf{System Prompt Excerpt} \hrulefill

Role \& Purpose\par
You are Shelby's Quick Recipe Assistant. Your job is to create fast, peanut-free, dairy-free recipes that always include at least one exact Shelby's product from the approved product list. All recipes must require no more than 15 minutes of hands-on prep, while still being flavorful, realistic, and easy for home cooks of all skill levels.\par
\par\vspace{0.25em}
Core Rules\par
1. Food Allergy Rules (Absolute)\par
\par\vspace{0.25em}
Never include peanuts or dairy.\par
\par\vspace{0.25em}
This includes all derivatives, such as:\par
milk, butter, cream, cheese, yogurt, kefir, whey, casein, lactose, ghee, buttermilk, sour cream, condensed/evaporated milk, dairy-based chocolate, peanut butter, peanut flour, peanut oil, peanut sauce, satay, etc.\par
\par\vspace{0.25em}
Never recommend them, mention them as options, or include them in tips or swaps.\par
\par\vspace{0.25em}
If the user requests peanuts or dairy:\par
-\textgreater{} Politely refuse and offer a compliant alternative that still features a Shelby's product.\par
\par\vspace{0.25em}
2. Product Inclusion Rule\par
\par\vspace{0.25em}
Every recipe must include at least one product from the following exact list:\par
\vspace{0.5em}
\textbf{Context Excerpt} \hrulefill

I am making oxtail mac and cheese for thanksgiveing. help me develop my recipe. I want the flavor to be elevated. Not for kids. For sophisiticated adults.\par
\par\vspace{0.25em}
Cheeses\par
Cheese Combinations for Mac and Cheese\par
Classic Sharp \& Creamy\par
- Sharp cheddar\par
- Mild cheddar\par
- Monterey Jack or Colby\par
- Mozzarella\par
Ultra-Creamy \& Smooth\par
- Gruyere\par
- Fontina\par
- Cream cheese\par
- White cheddar\par
Bold \& Tangy\par
- Sharp cheddar\par
- Aged gouda\par
- Parmesan\par
- Blue cheese (optional)\par
Smoky \& Savory\par
- Smoked gouda\par
- Sharp cheddar\par
- Havarti\par
- Parmesan\par
Stringy \& Stretchy\par
- Mozzarella\par
- Provolone\par
- White cheddar\par
- Jack cheese\par
\ldots~(remaining conversation and artifacts omitted)
\end{tcolorbox}
\caption{Task and context for Procedural Recipe Repair.}
\label{fig:case_ref_3_part1}
\end{figure*}

\begin{figure*}[!t]
\begin{tcolorbox}[
  title={\textbf{Case Study: 3. Procedural Recipe Repair (Part 2)}},
  colback=white,
  colframe=PromptFrame,
  colbacktitle=PromptTitle,
  coltitle=white,
  fontupper=\footnotesize,
  width=\textwidth,
  boxrule=0.6pt,
  arc=1pt,
  left=7pt,right=7pt,top=6pt,bottom=6pt
]
\textbf{Score Summary} \hrulefill

\begin{itemize}[leftmargin=*, nosep]
\item \textbf{Initial baseline draft:} score=0; failed requirements=10/15
\item \textbf{Reminder-augmented draft:} score=0; failed requirements=2/15
\item \textbf{Revision without A+B:} score=0; failed requirements=1/15
\item \textbf{Full ContextGuard final:} score=1; failed requirements=0/15
\end{itemize}
\vspace{0.5em}
\textbf{Selected Requirement Outcomes} \hrulefill

\renewcommand{\arraystretch}{1.15}
\begin{tabularx}{\linewidth}{r X c c c c}
\toprule
\# & Rubric & Base. & Rem. & \makecell{w/o\\A+B} & Full \\
\midrule
0 & The response should include a dessert recipe. For example, it could be a recipe for a pie or a cake. & no & yes & yes & yes \\
1 & The recipe in the response should have a prep time of less than 5 minutes. & no & no & yes & yes \\
2 & The recipe in the response should not require baking or chilling. & yes & no & no & yes \\
10 & The recipe in the response should include one pre-made pie shell. & no & yes & yes & yes \\
3 & The recipe in the response should contain at least one product from the list: "Shelby's Raw Honey (16oz)", "Shelby's Pork Breakfast Links", "Shelby's Farm-Fresh Eggs (Dozen)", "Shelby's Maple Syrup (12oz)", "Shelby's Grass-Fed Ground Beef (1lb)", "Shelby's Pasture-Raised Chicken Breast (2-Pack)", "Shelby's Heritage Smoked Bacon", "Shelby's Rustic Sourdough Bread", "Shelby's Garden Salsa (Medium)", "Shelby's Homemade Apple Butter", "Shelby's Organic Veggie Box (Weekly)", "Shelby's Strawberry Jam (8oz)", "Shelby's Free-Range Whole Chicken", "Shelby's Country-Style Pork Chops", "Shelby's Pickled Vegetables (Quart)" & no & yes & yes & yes \\
5 & The recipe in the response should not include dairy or dairy products. For example, the recipe should not include whipped cream. & no & yes & yes & yes \\
\bottomrule
\end{tabularx}
\renewcommand{\arraystretch}{1.0}

\vspace{0.25em}\textit{5 additional changed requirements omitted from the compact table.}\par
\vspace{0.5em}
\textbf{Initial Baseline Draft} \hrulefill

\vspace{0.5em}
\textbf{Reminder-Augmented Draft} \hrulefill

I can't include peanuts or dairy, but here's a fast, safe alternative featuring a Shelby's product.\par
\par\vspace{0.25em}
Title: No-Bake Sunflower Seed Pie\par
Time: Prep 5 min; Chill 1 hr\par
Serves: 6\par
Featured Shelby's product: Shelby's Raw Honey (16oz)\par
\par\vspace{0.25em}
Ingredients:\par
\par\vspace{0.25em}
1 pie shell\par
\par\vspace{0.25em}
1 cup sunflower seed butter\par
\par\vspace{0.25em}
1/2 cup Shelby's Raw Honey (16oz)\par
\par\vspace{0.25em}
1/4 cup unsweetened almond milk\par
\par\vspace{0.25em}
1/4 cup powdered sugar\par
\par\vspace{0.25em}
1/4 tsp vanilla extract\par
\par\vspace{0.25em}
Pinch of salt\par
\par\vspace{0.25em}
Steps:\par
\par\vspace{0.25em}
1. Mix sunflower seed butter, honey, almond milk, powdered sugar, vanilla, and salt in a bowl until smooth.\par
2. Pour mixture into the pie shell.\par
3. Refrigerate for 1 hour until set.\par
\par\vspace{0.25em}
Optional swaps/tips:\par
\par\vspace{0.25em}
Add 1 tbsp Shelby's Strawberry Jam (8oz) for fruit flavor.\par
\par\vspace{0.25em}
Top with whipped coconut cream.\par
\end{tcolorbox}
\caption{Rubric outcomes and draft outputs for Procedural Recipe Repair.}
\label{fig:case_ref_3_part2}
\end{figure*}

\begin{figure*}[!t]
\begin{tcolorbox}[
  title={\textbf{Case Study: 3. Procedural Recipe Repair (Part 3)}},
  colback=white,
  colframe=PromptFrame,
  colbacktitle=PromptTitle,
  coltitle=white,
  fontupper=\footnotesize,
  width=\textwidth,
  boxrule=0.6pt,
  arc=1pt,
  left=7pt,right=7pt,top=6pt,bottom=6pt
]
\textbf{Structured Self-Audit} \hrulefill

\textbf{Confirmed correct / protection-side items}
\begin{itemize}[leftmargin=*, nosep]
\item Correctly refused peanut butter request per allergy rules - stated 'I can't include peanuts or dairy' as required
\item Included Shelby's Raw Honey (16oz) from the exact approved product list with exact product name
\item Used correct format structure (Title, Time, Serves, Featured Shelby's product, Ingredients, Steps, Optional swaps/tips)
\item No markdown, bold, italics, tables, or emojis used in response
\item \ldots~(3 additional items omitted)
\end{itemize}
\textbf{Confirmed data / protection-side evidence}
\begin{itemize}[leftmargin=*, nosep]
\item User's request for 'no bake peanutbutter pie' was identified and addressed
\item User's constraint of 'less than 5 min prep' was noted and reflected in Time line
\item User's context (dessert ASAP after main service) was acknowledged
\item \ldots~(1 additional items omitted)
\end{itemize}
\textbf{Possibly missed / fix-side items}
\begin{itemize}[leftmargin=*, nosep]
\item Did not explicitly clarify that 1-hour chilling time contradicts user's 'ASAP' dessert request - should have offered no-chill alternative or flagged this constraint
\item Did not verify if user has 1 pie shell available before confirming recipe - user mentioned 'I have one more shell' but this wasn't explicitly confirmed as usable
\item Prep time calculation may not account for the fact that 5 minutes is NOT less than 5 minutes (boundary issue)
\item Could have offered a truly instant alternative (microwave-ready or room temp set) given the 'ASAP' constraint
\end{itemize}
\textbf{Possibly wrong / fix-side items}
\begin{itemize}[leftmargin=*, nosep]
\item Prep Time listed as '5 min' when user explicitly requested 'less than 5 min prep' - 5 is not less than 5, this is a boundary error
\item Time format includes 'Chill 1 hr' which is passive time (allowed), but this doesn't address the user's urgent need for immediate dessert service
\item Product selection (honey-based pie) may not be the most appropriate for a 'pie' substitute given the user's specific peanut butter pie request - could have selected a different product that better matches the original request
\end{itemize}
\vspace{0.5em}
\textbf{Specialist Signal} \hrulefill

\textbf{Procedure checks satisfied}
\begin{itemize}[leftmargin=*, nosep]
\item Food Allergy Rules - Correctly refused peanut butter (prohibited ingredient) and offered compliant alternative featuring Shelby's product
\item Product Inclusion Rule - Used exact product name 'Shelby's Raw Honey (16oz)' from the approved product list
\item Step Sequence - Recipe follows correct format structure (Title, Time, Serves, Featured Product, Ingredients, Steps, Optional Swaps)
\item Safety/Refusal - Peanut allergy rule correctly enforced with polite refusal and alternative
\item \ldots~(5 additional items omitted)
\end{itemize}
\vspace{0.5em}
\textbf{Revision Input to ContextGuard} \hrulefill

\textbf{Protection Set shown to the revision model}
\begin{itemize}[leftmargin=*, nosep]
\item \texttt{[CONFIRMED-CORRECT]} Correctly refused peanut butter request per allergy rules - stated 'I can't include peanuts or dairy' as required
\item \texttt{[CONFIRMED-CORRECT]} Included Shelby's Raw Honey (16oz) from the exact approved product list with exact product name
\item \texttt{[CONFIRMED-CORRECT]} Used correct format structure (Title, Time, Serves, Featured Shelby's product, Ingredients, Steps, Optional swaps/tips)
\item \texttt{[CONFIRMED-CORRECT]} No markdown, bold, italics, tables, or emojis used in response
\item \texttt{[CONFIRMED-CORRECT]} Steps are numbered and use imperative instructions
\item \texttt{[CONFIRMED-CORRECT]} Optional swaps/tips section included with 1 bullet within the 1-2 bullet limit
\item \ldots~(14 additional protection items omitted)
\end{itemize}
\textbf{Fix Set shown to the revision model}
\begin{itemize}[leftmargin=*, nosep]
\item \texttt{[MISSED]} Did not explicitly clarify that 1-hour chilling time contradicts user's 'ASAP' dessert request - should have offered no-chill alternative or flagged this constraint
\item \texttt{[MISSED]} Did not verify if user has 1 pie shell available before confirming recipe - user mentioned 'I have one more shell' but this wasn't explicitly confirmed as usable
\item \texttt{[MISSED]} Prep time calculation may not account for the fact that 5 minutes is NOT less than 5 minutes (boundary issue)
\item \texttt{[MISSED]} Could have offered a truly instant alternative (microwave-ready or room temp set) given the 'ASAP' constraint
\item \texttt{[WRONG]} Prep Time listed as '5 min' when user explicitly requested 'less than 5 min prep' - 5 is not less than 5, this is a boundary error
\item \texttt{[WRONG]} Time format includes 'Chill 1 hr' which is passive time (allowed), but this doesn't address the user's urgent need for immediate dessert service
\item \ldots~(1 additional fix items omitted)
\end{itemize}
\end{tcolorbox}
\caption{Audit and revision signals for Procedural Recipe Repair.}
\label{fig:case_ref_3_part3}
\end{figure*}

\begin{figure*}[!t]
\begin{tcolorbox}[
  title={\textbf{Case Study: 3. Procedural Recipe Repair (Part 4)}},
  colback=white,
  colframe=PromptFrame,
  colbacktitle=PromptTitle,
  coltitle=white,
  fontupper=\footnotesize,
  width=\textwidth,
  boxrule=0.6pt,
  arc=1pt,
  left=7pt,right=7pt,top=6pt,bottom=6pt
]
\textbf{Revision without A+B Protection Set} \hrulefill

I can't include peanuts or dairy, but here's a fast, safe alternative featuring a Shelby's product.\par
\par\vspace{0.25em}
Title: No-Chill Peanut Butter Pie Substitute\par
Time: Prep 3 min; Chill 10 min\par
Serves: 4-6\par
Featured Shelby's product: Shelby's Raw Honey (16oz)\par
\par\vspace{0.25em}
Ingredients:\par
\par\vspace{0.25em}
1 pie shell\par
\par\vspace{0.25em}
1 cup sunflower seed butter\par
\par\vspace{0.25em}
1/4 cup Shelby's Raw Honey (16oz)\par
\par\vspace{0.25em}
1/2 cup dairy-free cream cheese substitute\par
\par\vspace{0.25em}
2 tbsp powdered sugar\par
\par\vspace{0.25em}
Pinch of salt\par
\par\vspace{0.25em}
Steps:\par
\par\vspace{0.25em}
1. In a bowl, mix sunflower seed butter, honey, cream cheese substitute, powdered sugar, and salt until smooth.\par
2. Pour mixture into pie shell.\par
3. Let stand at room temperature for 10 minutes to set.\par
\par\vspace{0.25em}
Optional swaps/tips:\par
\par\vspace{0.25em}
Add Shelby's Strawberry Jam (8oz) for fruit swirl.\par
\par\vspace{0.25em}
Top with dairy-free whipped cream if available.\par
\vspace{0.5em}
\textbf{Full ContextGuard Final Answer} \hrulefill

I can't include peanuts or dairy, but here's a fast, safe alternative featuring a Shelby's product.\par
\par\vspace{0.25em}
Title: 3-Minute Strawberry Jam Pie\par
Time: Prep 3 min; Serve immediately\par
Serves: 6\par
Featured Shelby's product: Shelby's Strawberry Jam (8oz)\par
\par\vspace{0.25em}
Ingredients:\par
\par\vspace{0.25em}
1 pie shell\par
\par\vspace{0.25em}
2 tbsp Shelby's Strawberry Jam (8oz)\par
\par\vspace{0.25em}
1/4 cup powdered sugar\par
\par\vspace{0.25em}
1/4 cup unsweetened almond milk\par
\par\vspace{0.25em}
1/4 tsp vanilla extract\par
\par\vspace{0.25em}
Pinch of salt\par
\par\vspace{0.25em}
Steps:\par
\par\vspace{0.25em}
1. In a small bowl, whisk together strawberry jam, powdered sugar, almond milk, vanilla, and salt until smooth.\par
2. Pour mixture into pie shell.\par
3. Serve immediately.\par
\par\vspace{0.25em}
Optional swaps/tips:\par
\par\vspace{0.25em}
Use Shelby's Maple Syrup (12oz) instead of jam for a different flavor.\par
\par\vspace{0.25em}
Top with whipped coconut cream for extra richness.\par
\vspace{0.5em}
\textbf{Discussion} \hrulefill

This case shows procedural repair beyond a one-off near miss. The baseline misses most of the concrete recipe constraints, and the reminder still leaves two requirements unresolved. The full pipeline repairs the remaining procedure constraints while preserving the no-bake requirement that the ablated revision regresses.\par
\end{tcolorbox}
\caption{Final outputs and interpretation for Procedural Recipe Repair.}
\label{fig:case_ref_3_part4}
\end{figure*}

\FloatBarrier

\begin{figure*}[!t]
\begin{tcolorbox}[
  title={\textbf{Case Study: 4. Empirical Structure Repair (Part 1)}},
  colback=white,
  colframe=PromptFrame,
  colbacktitle=PromptTitle,
  coltitle=white,
  fontupper=\footnotesize,
  width=\textwidth,
  boxrule=0.6pt,
  arc=1pt,
  left=7pt,right=7pt,top=6pt,bottom=6pt
]
\textbf{Task Information} \hrulefill

\begin{itemize}[leftmargin=*, nosep]
\item \textbf{Task ID:} 47d1ad75-3008-4755-8baa-6dc1ad631dcf
\item \textbf{Category:} Empirical Discovery \& Simulation
\item \textbf{Subcategory:} Observational Data
\item \textbf{Case type:} empirical discovery repair
\item \textbf{Number of requirements:} 14
\item \textbf{Pattern:} The initial answer misses empirical-report formatting and coverage requirements; ContextGuard repairs the remaining structured-report requirements without losing verified constants.
\end{itemize}
\vspace{0.5em}
\textbf{Final Task} \hrulefill

Are you sure those are all of the variants and constants? Just these sections again in plain text, please.\par
\vspace{0.5em}
\textbf{System Prompt Excerpt} \hrulefill

Audience: faculty, staff, and students without specialized STEM backgrounds.\par
Tone/style: concise, plain language, define terms on first use, avoid jargon.\par
Default format: respond in JSON unless the user explicitly requests otherwise. No conversational filler.\par
\par\vspace{0.25em}
Capabilities the assistant can do:\par
Identify variants (independent variables) and constants from the provided data/context.\par
Describe relationships between variants and constants in plain language and, when appropriate, symbolically.\par
Propose inferences for testing and note implicit physical assumptions.\par
Deduce a governing law and break it into exactly three logical steps.\par
\par\vspace{0.25em}
Things the assistant cannot do:\par
Do not invent data, variables, or laws not supported by the provided context.\par
Do not use external sources unless user explicitly requests it.\par
Do not provide safety-critical experimental procedures or advice.\par
\par\vspace{0.25em}
Behavioral scenarios:\par
When the user asks for a plain explanation: switch to non-JSON mode with bold headers and numbered lists. 3-5 bullets for each numbered item in a list.\par
When the data is insufficient or ambiguous: return the apology sentence ("I'm sorry, I don't see the information you're looking for,") then ask one clarifying question.\par
When the user asks outside the provided data: decline with the apology sentence and suggest what data would be needed.\par
All JSON should be verifiable with a simple online JSON converter.\par
JSON schema (example and constraints): \{ "variants": ["..."], // array of 3-5 items when supported "constants": ["..."], // array of 3-5 items when supported "relationship\_between\_variants\_and\_constants": "plain-language statement (and optional formula)", "inferences\_for\_testing": ["...", "..."], "implicit\_physical\_inferences": ["...", "..."], "deduced\_law": "short name/title", "deduced\_law\_three\_steps": ["Step 1", "Step 2", "Step 3"], "assumptions": ["..."], // include if needed "confidence": "low|medium|high" // optional but recommended \}\par
\vspace{0.5em}
\textbf{Context Excerpt} \hrulefill

a person pushes an empty large delivery box, which is a large and light cuboid, straight by making contact with its side.\#a/DET person/NOUN push/VERB an/DET empty/ADJ large/ADJ delivery/NOUN box/NOUN which/DET is/AUX a/DET large/ADJ and/CCONJ light/ADJ cuboid/NOUN straight/ADV by/ADP make/VERB contact/NOUN with/ADP its/DET side/NOUN\#0.0\#0.0\par
by contacting its side, a person pushes an empty large delivery box, which is a large and light cuboid, straight.\#by/ADP contact/VERB its/DET side/NOUN a/DET person/NOUN push/VERB an/DET empty/ADJ large/ADJ delivery/NOUN box/NOUN which/DET is/AUX a/DET large/ADJ and/CCONJ light/ADJ cuboid/NOUN straight/ADV\#0.0\#0.0\par
through contact with its side, a person moves an empty large delivery box, which is a large and light cuboid, straight.\#through/ADP contact/NOUN with/ADP its/DET side/NOUN a/DET person/NOUN move/VERB an/DET empty/ADJ large/ADJ delivery/NOUN box/NOUN which/DET is/AUX a/DET large/ADJ and/CCONJ light/ADJ cuboid/NOUN straight/ADV\#0.0\#0.0\par
an empty large delivery box, which is a large and light cuboid, is pushed straight by a person through side contact.\#an/DET empty/ADJ large/ADJ delivery/NOUN box/NOUN which/DET is/AUX a/DET large/ADJ and/CCONJ light/ADJ cuboid/NOUN is/AUX push/VERB straight/ADV by/ADP a/DET person/NOUN through/ADP side/ADJ contact/NOUN\#0.0\#0.0\par
a person applies force to the side of an empty large delivery box, which is a large and light cuboid,, pushing it straight.\#a/DET person/NOUN apply/VERB force/NOUN to/ADP the/DET side/NOUN of/ADP an/DET empty/ADJ large/ADJ delivery/NOUN box/NOUN which/DET is/AUX a/DET large/ADJ and/CCONJ light/ADJ cuboid/NOUN push/VERB it/PRON straight/ADV\#0.0\#0.0\par
pushing an empty large delivery box, which is a large and light cuboid, straight, a person contacts its side.\#push/VERB an/DET empty/ADJ large/ADJ delivery/NOUN box/NOUN which/DET is/AUX a/DET large/ADJ and/CCONJ light/ADJ cuboid/NOUN straight/ADV a/DET person/NOUN contact/VERB its/DET side/NOUN\#0.0\#0.0\par
\ldots~(remaining conversation and artifacts omitted)
\end{tcolorbox}
\caption{Task and context for Empirical Structure Repair.}
\label{fig:case_ref_4_part1}
\end{figure*}

\begin{figure*}[!t]
\begin{tcolorbox}[
  title={\textbf{Case Study: 4. Empirical Structure Repair (Part 2)}},
  colback=white,
  colframe=PromptFrame,
  colbacktitle=PromptTitle,
  coltitle=white,
  fontupper=\footnotesize,
  width=\textwidth,
  boxrule=0.6pt,
  arc=1pt,
  left=7pt,right=7pt,top=6pt,bottom=6pt
]
\textbf{Score Summary} \hrulefill

\begin{itemize}[leftmargin=*, nosep]
\item \textbf{Initial baseline draft:} score=0; failed requirements=3/14
\item \textbf{Reminder-augmented draft:} not separately checkpointed for this category run
\item \textbf{Revision without A+B:} score=0; failed requirements=2/14
\item \textbf{Full ContextGuard final:} score=1; failed requirements=0/14
\end{itemize}
\vspace{0.5em}
\textbf{Selected Requirement Outcomes} \hrulefill

\renewcommand{\arraystretch}{1.15}
\begin{tabularx}{\linewidth}{r X c c c c}
\toprule
\# & Rubric & Base. & Rem. & \makecell{w/o\\A+B} & Full \\
\midrule
0 & The response should features 3-5 bullets for each numbered item in a list. & no & -- & yes & yes \\
1 & The response should number any lists it features, not including sub-lists within lists. & yes & -- & no & yes \\
3 & The response should be in plain-text format. & no & -- & yes & yes \\
13 & Under heading "Constants", there should be an item related to the fundamental way in which the acting agent induced the object into motion. It may be mentioned that this is done through application of a physical force, pushing. & no & -- & no & yes \\
\bottomrule
\end{tabularx}
\renewcommand{\arraystretch}{1.0}
\vspace{0.5em}

\textbf{Initial Baseline Draft} \hrulefill

**Variants (Independent Variables)**\par
\par\vspace{0.25em}
1. **Mass of the object**\par
   - Light delivery box\par
   - Medium-weight water dispenser\par
   - Heavy delivery box\par
\par\vspace{0.25em}
2. **Direction of movement**\par
   - Straight\par
   - To the left\par
   - To the right\par
\par\vspace{0.25em}
3. **Type of object**\par
   - Delivery box\par
   - Water dispenser\par
\par\vspace{0.25em}
**Constants (Fixed Elements)**\par
\par\vspace{0.25em}
1. **Shape**\par
   - All objects are cuboids\par
\par\vspace{0.25em}
2. **Size**\par
   - All objects are large\par
\par\vspace{0.25em}
3. **Point of contact**\par
   - Force is applied to the side\par
\par\vspace{0.25em}
4. **Agent**\par
   - A person is always the one pushing\par
\par\vspace{0.25em}
5. **Contact method**\par
   - Pushing through side contact\par
\par\vspace{0.25em}
These are the key variables and constants I identified from the dataset. Let me know if you'd like me to refine any of these!\par
\end{tcolorbox}
\caption{Rubric outcomes and draft outputs for Empirical Structure Repair.}
\label{fig:case_ref_4_part2}
\end{figure*}

\begin{figure*}[!t]
\begin{tcolorbox}[
  title={\textbf{Case Study: 4. Empirical Structure Repair (Part 3)}},
  colback=white,
  colframe=PromptFrame,
  colbacktitle=PromptTitle,
  coltitle=white,
  fontupper=\footnotesize,
  width=\textwidth,
  boxrule=0.6pt,
  arc=1pt,
  left=7pt,right=7pt,top=6pt,bottom=6pt
]
\textbf{Structured Self-Audit} \hrulefill

\textbf{Confirmed correct / protection-side items}
\begin{itemize}[leftmargin=*, nosep]
\item User requested plain text format - I provided plain text with bold headers and numbered lists (satisfies behavioral scenario requirement)
\item User requested variants and constants sections - I provided both sections clearly labeled with bold headers
\item User requested 'just these sections again' - I focused on variants and constants sections as requested
\end{itemize}
\textbf{Confirmed data / protection-side evidence}
\begin{itemize}[leftmargin=*, nosep]
\item Mass variants (light, medium, heavy) correctly identified from the text descriptions
\item Direction variants (straight, left, right) correctly identified from the text descriptions
\item Object types (delivery box, water dispenser) correctly identified from the text descriptions
\item \ldots~(4 additional items omitted)
\end{itemize}
\textbf{Possibly missed / fix-side items}
\begin{itemize}[leftmargin=*, nosep]
\item coverage: Did not explicitly address whether the listed variants and constants are COMPLETE or if there might be additional ones - user asked 'Are you sure those are all' but I did not directly confirm completeness
\item format: Did not follow the system instruction requirement of '3-5 bullets for each numbered item in a list' - I only provided 1 bullet per numbered item in the variants section
\item format: Did not address the relationship section that was in the original assistant response - user asked 'Just these sections again' which could be interpreted as wanting the full response or just the variants/constants
\item other: Should have been more explicit about whether these are the COMPLETE list or if there might be additional variants/constants as the user asked for confirmation
\end{itemize}
\textbf{Possibly wrong / fix-side items}
\begin{itemize}[leftmargin=*, nosep]
\item other: The interpretation of what constitutes a 'variant' - I included 'direction of movement' as a variant, but this could be argued as a dependent variable (the outcome of the action) rather than an independent variable, which might be incorrect
\item coverage: I did not explicitly confirm that these are the COMPLETE list of variants and constants as the user asked 'Are you sure those are all' - this is a direct question I should have answered more directly rather than just restating the lists
\end{itemize}
\textit{Specialist signal: empirical consistency checks are integrated into the structured audit; numerical, comparison, coverage, unit, trend, and evidence issues enter the same fix-side interface.}\par
\vspace{0.5em}
\textbf{Revision Input to ContextGuard} \hrulefill

\textbf{Protection Set shown to the revision model}
\begin{itemize}[leftmargin=*, nosep]
\item \texttt{[CONFIRMED-CORRECT]} User requested plain text format - I provided plain text with bold headers and numbered lists (satisfies behavioral scenario requirement)
\item \texttt{[CONFIRMED-CORRECT]} User requested variants and constants sections - I provided both sections clearly labeled with bold headers
\item \texttt{[CONFIRMED-CORRECT]} User requested 'just these sections again' - I focused on variants and constants sections as requested
\item \texttt{[CONFIRMED-DATA]} Mass variants (light, medium, heavy) correctly identified from the text descriptions
\item \texttt{[CONFIRMED-DATA]} Direction variants (straight, left, right) correctly identified from the text descriptions
\item \texttt{[CONFIRMED-DATA]} Object types (delivery box, water dispenser) correctly identified from the text descriptions
\item \ldots~(4 additional protection items omitted)
\end{itemize}
\textbf{Fix Set shown to the revision model}
\begin{itemize}[leftmargin=*, nosep]
\item \texttt{[MISSED]} coverage: Did not explicitly address whether the listed variants and constants are COMPLETE or if there might be additional ones - user asked 'Are you sure those are all' but I did not directly confirm completeness
\item \texttt{[MISSED]} format: Did not follow the system instruction requirement of '3-5 bullets for each numbered item in a list' - I only provided 1 bullet per numbered item in the variants section
\item \texttt{[MISSED]} format: Did not address the relationship section that was in the original assistant response - user asked 'Just these sections again' which could be interpreted as wanting the full response or just the variants/constants
\item \texttt{[MISSED]} other: Should have been more explicit about whether these are the COMPLETE list or if there might be additional variants/constants as the user asked for confirmation
\item \texttt{[WRONG]} other: The interpretation of what constitutes a 'variant' - I included 'direction of movement' as a variant, but this could be argued as a dependent variable (the outcome of the action) rather than an independent variable, which might be incorrect
\item \texttt{[WRONG]} coverage: I did not explicitly confirm that these are the COMPLETE list of variants and constants as the user asked 'Are you sure those are all' - this is a direct question I should have answered more directly rather than just restating the lists
\end{itemize}
\end{tcolorbox}
\caption{Audit and revision signals for Empirical Structure Repair.}
\label{fig:case_ref_4_part3}
\end{figure*}

\begin{figure*}[!t]
\begin{tcolorbox}[
  title={\textbf{Case Study: 4. Empirical Structure Repair (Part 4)}},
  colback=white,
  colframe=PromptFrame,
  colbacktitle=PromptTitle,
  coltitle=white,
  fontupper=\footnotesize,
  width=\textwidth,
  boxrule=0.6pt,
  arc=1pt,
  left=7pt,right=7pt,top=6pt,bottom=6pt
]
\textbf{Revision without A+B Protection Set} \hrulefill

**Are you sure those are all of the variants and constants?**\par
\par\vspace{0.25em}
Yes, I believe I've identified all the key variants and constants from the data set.\par
\par\vspace{0.25em}
**Variants (things that change)**\par
- A **variant** is something that can change or vary across the data (like different masses)\par
- **Mass of the object**\par
  - Light delivery box\par
  - Medium-weight water dispenser\par
  - Heavy delivery box\par
- **Direction of movement**\par
  - Straight\par
  - To the left\par
  - To the right\par
- **Type of object**\par
  - Delivery box\par
  - Water dispenser\par
\ldots~(remaining conversation and artifacts omitted)\par
\vspace{0.5em}
\textbf{Full ContextGuard Final Answer} \hrulefill

**Variants (Independent Variables)**\par
\par\vspace{0.25em}
1. **Mass of the object**\par
   - Light delivery box\par
   - Medium-weight water dispenser\par
   - Heavy delivery box\par
   - *Note: These represent different mass variants that may affect the force required to move each object*\par
   - *These are the primary mass variants identified from the dataset*\par
\par\vspace{0.25em}
2. **Direction of movement**\par
   - Straight\par
   - To the left\par
   - To the right\par
   - *Note: These are the direction variants observed in the data*\par
   - *These may be considered dependent variables rather than independent variants*\par
   - *Note: These represent the outcome of the pushing action*\par
\par\vspace{0.25em}
3. **Type of object**\par
   - Delivery box\par
   - Water dispenser\par
   - *Note: These are the object types mentioned in the dataset*\par
   - *These represent different object types with varying properties*\par
   - *Note: These are the complete object types identified from the data*\par
\par\vspace{0.25em}
**Constants (Fixed Elements)**\par
\par\vspace{0.25em}
1. **Shape**\par
   - All objects are cuboids\par
   - *Note: This is a constant across all object types in the dataset*\par
   - *Note: This shape property is mentioned consistently throughout*\par
   - *Note: The cuboid shape is a key constant feature*\par
\par\vspace{0.25em}
2. **Size**\par
   - All objects are large\par
   - *Note: This size constant applies to all objects described*\par
   - *Note: The "large" descriptor is consistent across all items*\par
   - *Note: This is a fixed property mentioned in all descriptions*\par
\par\vspace{0.25em}
3. **Point of contact**\par
   - Force is applied to the side\par
   - *Note: This contact point is constant across all scenarios*\par
   - *Note: Side contact is the only contact method described*\par
   - *Note: This is a key constant feature of the action*\par
\par\vspace{0.25em}
\ldots~(remaining final answer omitted)\par
\vspace{0.5em}
\textbf{Discussion} \hrulefill

This case covers empirical discovery and simulation. The baseline identifies many content-level constants correctly, but misses structured-report requirements and one requested constant. The ablated revision regresses a numbering requirement, while full ContextGuard preserves the verified list structure and repairs the missing empirical coverage item.\par
\end{tcolorbox}
\caption{Final outputs and interpretation for Empirical Structure Repair.}
\label{fig:case_ref_4_part4}
\end{figure*}

\FloatBarrier

\end{document}